\documentclass[11pt,a4paper,fleqn]{extarticle}

\usepackage[
    top=20mm,
    bottom=20mm,
    left=18mm,
    right=18mm
]{geometry}

\usepackage{graphicx}
\usepackage{titlesec}
\usepackage{fancyhdr}
\usepackage{array}
\usepackage{booktabs}
\usepackage{tikz}
\usepackage[most]{tcolorbox}
\usepackage{amssymb}
\usepackage{amsmath}
\usepackage{fontspec}
\usepackage{xcolor}
\usepackage[normalem]{ulem}
\usepackage{environ}
\usepackage{etoolbox}

\usepackage[numbers,compress]{natbib}
\usepackage{nicefrac}
\usepackage{microtype}
\usepackage{adjustbox}
\usepackage{float}
\usepackage{placeins}
\usepackage{subcaption}
\usepackage{multirow}
\usepackage{makecell}
\usepackage{enumitem}
\usepackage{xspace}
\usepackage{algorithm}
\usepackage{algpseudocode}
\usepackage{listings}
\usepackage{etoc}
\usepackage{caption}
\usepackage{hyperref}
\usepackage{url}

\titleformat{\section}
    {\sffamily\large\bfseries}
    {\thesection}
    {1em}
    {}

\titleformat{\subsection}
  {\sffamily\normalsize\bfseries}
  {\thesubsection}
  {1em}
  {}

\titleformat{\subsubsection}
  {\sffamily\normalsize\bfseries}
  {\thesubsubsection}
  {1em}
  {}

\titleformat{\paragraph}[runin]
  {\normalsize\bfseries}
  {\theparagraph}
  {1em}
  {}

\etocsettocstyle{\section*{\contentsname}\setlength{\parskip}{0pt}}{}
\makeatletter
\renewcommand*\l@subsection{\@dottedtocline{2}{1.5em}{3.2em}}
\makeatother

\definecolor{tether-gray}{HTML}{F2F1EF}
\definecolor{tether-link}{HTML}{1A7F64}

\hypersetup{
    colorlinks=true,
    linkcolor=tether-link,
    citecolor=tether-link,
    urlcolor=tether-link
}

\newtcolorbox[auto counter]{custombox}[2][]{
    title={Box~\thetcbcounter: #2},
    label={#1},
    fonttitle=\bfseries,
    coltitle=black,
    before title=\vspace{15pt},
    colback=tether-gray,
    colframe=tether-gray,
    boxrule=0pt,
    arc=5pt,
    left=15pt,
    right=15pt,
    bottom=15pt,
    width=\textwidth,
}

\tcbset{tethertable/.style={
    colback=tether-gray,
    colframe=tether-gray,
    boxrule=0pt,
    arc=5pt,
    left=10pt,
    right=10pt,
    top=8pt,
    bottom=8pt,
    before skip=0pt,
    after skip=0pt
}}
\let\tetherorigtable\table
\let\endtetherorigtable\endtable
\renewenvironment{table}[1][tbp]
  {\tetherorigtable[#1]\begin{tcolorbox}[tethertable]}
  {\end{tcolorbox}\endtetherorigtable}

\NewEnviron{ack}{%
  \section*{Acknowledgments and Disclosure of Funding}
  \BODY
}

\newcommand{\medpsy}{\textsc{MedPsy}\xspace}
\newcommand{\mqfourb}{\textsc{MedPsy}-4B\xspace}
\newcommand{\mqonesevenb}{\textsc{MedPsy}-1.7B\xspace}
\newcommand{\bestcell}[1]{\textbf{#1}}

\newif\ifshowchanges
\showchangesfalse
\ifshowchanges
  
  \newcommand{\old}[1]{\textcolor{red!70!black}{\sout{#1}}}
  
\else
  
  \newcommand{\old}[1]{}
  
\fi

\newcommand{\papertitle}{MedPsy: State-of-the-Art Small Medical Language Models for Efficient Edge Deployment}

\newcommand{\paperabstract}{%
Medical language models promise expert clinical support, but the strongest open-source LLMs remain too large for private edge-device deployment, while compact medical models still trail substantially on knowledge-intensive and real-world healthcare tasks. Moreover, reasoning models that fit edge memory budgets can produce long reasoning traces, increasing sequential decoding, end-to-end latency, energy consumption, and KV-cache memory pressure. We present \medpsy, a family of text-only 1.7B and 4B medical language models designed for edge deployment, explicitly targeting this accuracy–generation-efficiency trade-off. Our recipe combines a synthetic medical data pipeline over biology, medicine, and health seeds with chain-of-thought targets from a 235B medical-focused LLM teacher; a four-stage post-training curriculum comprising two SFT stages followed by two RL stages with hard-sample mining; and an edge deployment evaluation spanning multiple GGUF quantization variants and real-device inference on laptop and smartphone hardware. On seven closed-ended medical benchmarks, \mqfourb{} surpasses MedGemma-1.5-4B by 19.34 points and matches MedGemma-27B while being 6.75× smaller; \mqonesevenb{} outperforms MedGemma-1.5-4B by 11.42 points despite being less than half its size. On HealthBench-Hard, \mqfourb{} surpasses MedGemma-27B by 15.33 points, while \mqonesevenb{} surpasses it by 11.66 points despite being ~16× smaller. Beyond accuracy, \medpsy reduces average response length by 1.7× (1.7B) and 3.2× (4B) versus its Qwen3 backbones, while \texttt{Q4\_K\_M} quantization preserves the aggregate score within 1 point of BF16 with $\sim69\%$ lower disk footprint. In real-device inference, \mqfourb{}’s shorter generations translate into $75\%$ lower median end-to-end latency and estimated energy per answer than its Qwen3 backbone at matched quantization, while peak RAM remains below 5 GB across our tested laptop and smartphone configurations.%
}

\newcommand{\huggingfaceurl}{https://huggingface.co/collections/qvac/medpsy}
\newcommand{\githuburl}{https://github.com/tether-ai-research/qvac-medpsy}

\newcommand{\paperlink}[3]{%
  \href{#2}{\raisebox{-0.2em}{\includegraphics[height=1.15em]{#1}}\hspace{0.4em}\textcolor{black}{\bfseries #3}}%
}

\begin{document}

\noindent
\includegraphics[height=20pt]{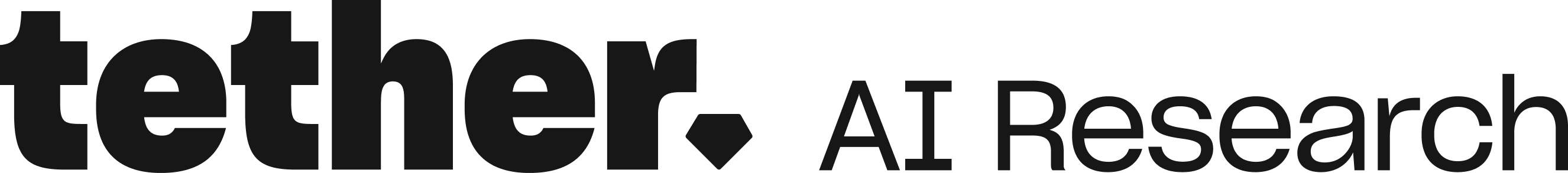}%
\hfill
\raisebox{0.5pt}{\includegraphics[height=14pt]{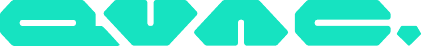}}
\par\medskip

\vspace{10pt}

\begin{tcolorbox}[
    colback=tether-gray,
    colframe=tether-gray,
    boxrule=0pt,
    arc=5pt,
    left=15pt,
    right=15pt,
    top=15pt,
    bottom=15pt,
    width=\textwidth
]

{\sffamily\fontsize{20.74}{25}\selectfont\bfseries\raggedright\hyphenpenalty=10000
\papertitle
\par}

\vspace{5pt}

{\sffamily\fontsize{11}{13.6}\selectfont\bfseries
Davide Vitabile,
Alexandro Buffa,
Akshay Nambiar,
Amril Nazir
\par}

\vspace{2.5pt}

{\sffamily\fontsize{11}{13.6}\selectfont
Tether Data, S.A. de C.V. d.b.a. Tether AI Research
\par}

\vspace{6pt}

{\fontsize{11}{13.6}\selectfont
\sffamily
\paperlink{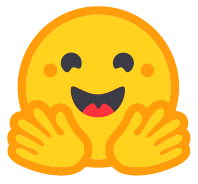}{\huggingfaceurl}{\mdseries\huggingfaceurl}\\[3pt]
\paperlink{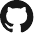}{\githuburl}{\mdseries\githuburl}
\par}

\vspace{15pt}

{\normalsize
\paperabstract
\par}

\end{tcolorbox}

\suppressfloats[t]

\vspace{10pt}

\section{Introduction}
\label{sec:intro}

\begin{figure}[t]
\centering
\begin{subfigure}[t]{0.495\linewidth}
  \centering
  \includegraphics[width=\linewidth]{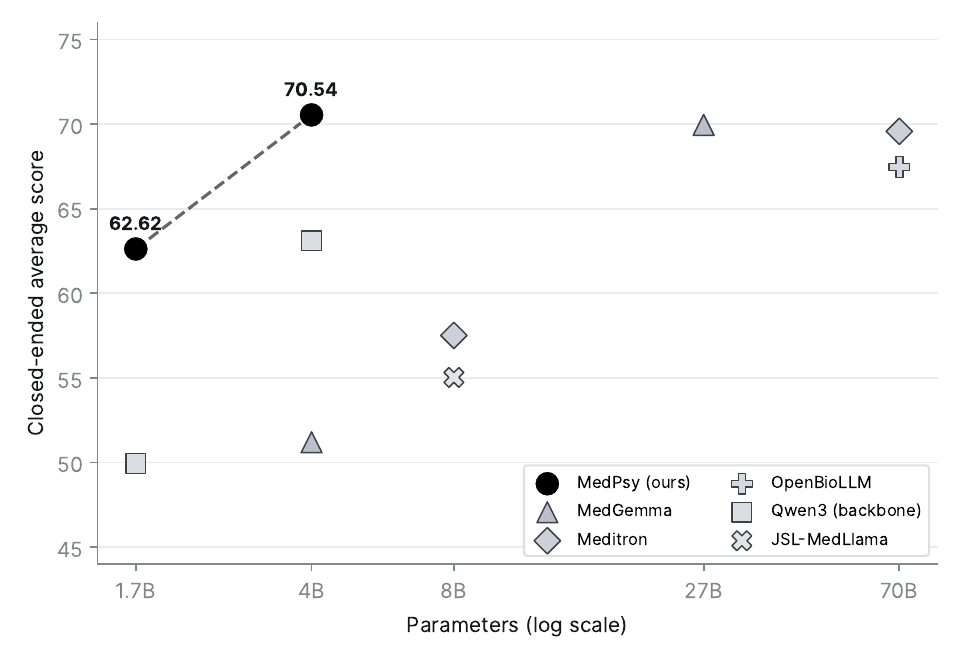}
  \caption{Capability per parameter: closed-ended average score vs.\ parameter count (log scale). \mqfourb{} matches 27B--70B medical LLMs at a fraction of the size; the dashed line connects the \medpsy{} pareto frontier. Scores from Tables~\ref{tab:closed} and~\ref{tab:large-medical-baselines}.}
  \label{fig:params-vs-score}
\end{subfigure}\hfill
\begin{subfigure}[t]{0.495\linewidth}
  \centering
  \includegraphics[width=\linewidth]{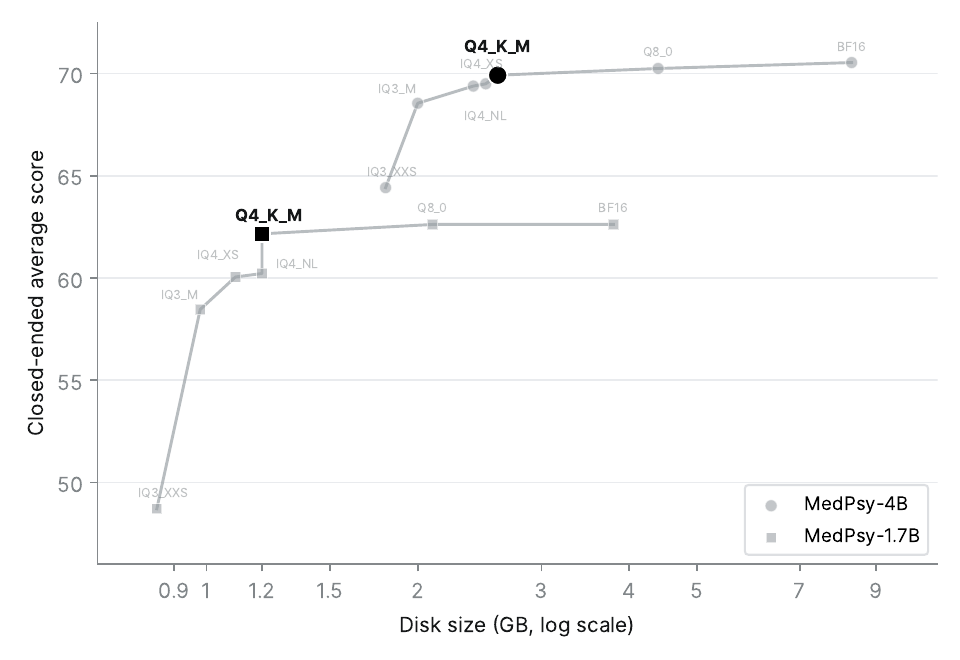}
  \caption{Capability per gigabyte: closed-ended average score vs.\ GGUF disk size (log scale) across seven quantization formats. Scores stay near BF16 until 3-bit precision, with a sharp cliff at \texttt{IQ3\_XXS}; highlighted points mark the capacity-vs-precision crossover. Data from Table~\ref{tab:q4-deployment}(a).}
  \label{fig:quant-sizes}
\end{subfigure}
\caption{\textbf{Efficiency along both deployment axes.} \medpsy{} models deliver frontier-level medical accuracy per parameter \textbf{(a)} and per gigabyte of quantized footprint \textbf{(b)}.}
\label{fig:efficiency}
\end{figure}

Medical knowledge and capabilities of language models have advanced rapidly, yet their deployment remains centralized. Among open-weight medical systems, strong models such as MedGemma-27B-text-it~\cite{medgemma2025} still require substantial accelerator memory for practical deployment, while proprietary medical systems such as Med-PaLM-style models and large GPT-4-based clinical assistants are typically accessed through cloud infrastructure. Both fundamentally conflict with the privacy, latency, and reliability requirements of healthcare. Patient records, symptom descriptions, and clinical notes are protected health information governed by strict regulatory regimes (HIPAA, GDPR, and equivalents). The dominant cloud paradigm requires this data to leave the device, traverse network infrastructure, and be processed remotely, creating attack surfaces, compliance burdens, and a fundamental tension between AI capability and patient privacy.

\paragraph{The compact-medical gap.} The current open landscape has a sharp trade-off between capability and deployability. MedGemma-27B-text-it delivers strong medical performance, but at 27B parameters it is infeasible for edge deployment, requiring tens of GB of VRAM. MedGemma-1.5-4B-it~\cite{medgemma15} is technically runnable on a high-end laptop but lacks the domain-specific capabilities of its larger variant on medical benchmarks, and, to the best of our knowledge, no open model in the 1--4B range delivers accuracy comparable to large medical-model baselines. This is not a pure compression problem: smaller models trained with conventional recipes suffer disproportionately on knowledge-intensive medical tasks, where factual precision and safe, well-structured responses are essential. Nor does fitting device memory guarantee efficient inference: compact reasoning models can still overthink, producing long traces that increase sequential decoding latency, energy use, and KV-cache traffic.

\paragraph{Our thesis.} \emph{For compact medical LLMs, medical capability and deployment efficiency can be optimized jointly: high-quality synthetic supervision and staged post-training build accuracy, while length-aware alignment and quantization reduce generation and memory costs for private on-device use.} We test this thesis by constructing the \medpsy{}\footnote{In MedPsy, ``Med'' denotes the medical specialization, while ``Psy'' is the shared suffix used across our organization's model families, inspired by psychohistory in Isaac Asimov's fiction rather than psychology.} family of two text-only medical models, \mqonesevenb{} and \mqfourb{}\footnote{Models and supporting artifacts are available at \url{https://huggingface.co/collections/qvac/medpsy}.}, on top of Qwen3 backbones~\cite{qwen3} with a four-stage post-training recipe (two Supervised Fine-Tuning (SFT) stages over a synthetic medical corpus built from public Genesis III biology/medicine seed prompts~\cite{genesis3} and newly generated health seeds derived from FineFineWeb~\cite{finefineweb}, then two Reinforcement Learning (RL) stages with hard-sample mining inspired by Fleming-R1~\cite{fleming-r1}), and we evaluate them across eight benchmark suites spanning clinical exams (MedQA-USMLE~\cite{medqa}, MedMCQA~\cite{medmcqa}), health literacy (MMLU-Health, MMLU-Pro Health~\cite{mmlu, mmlu-pro}), expert reasoning (MedXpertQA~\cite{medxpertqa}), biomedical research (PubMedQA~\cite{pubmedqa}), underserved-region medicine (AfriMedQA~\cite{afrimedqa}), and real-world clinical scenarios (HealthBench and HealthBench-Hard~\cite{healthbench}).

\paragraph{Contributions.} Our contributions are:
\begin{itemize}[leftmargin=*, itemsep=2pt, topsep=2pt]
\item \textbf{State-of-the-Art Efficient Models.} \mqfourb{} matches MedGemma-27B on closed-ended accuracy and surpasses on HealthBench suites both MedGemma-4B (+20/+28.33) and 27B  (+9.0/+15.3). Similarly, \mqonesevenb{} exceeds MedGemma-4B on closed-ended tasks and outperforms both MedGemma variants on HealthBench, achieving these leads at up to 16$\times$ smaller scale (Fig.~\ref{fig:params-vs-score}; Sec.~\ref{sec:eval}).
\item \textbf{Post-training recipe.} We combine single-teacher synthetic supervision, broad-to-focused SFT, and easy-to-hard DAPO with deterministic failure mining. Component ablations separate the roles of SFT, DAPO's drift-enabling changes, the overlong-response penalty, and Stage-2 mining (Sec.~\ref{subsec:stages}; App.~\ref{app:ablations}).
\item \textbf{Token efficiency for the edge.} \mqfourb{} and \mqonesevenb{} achieve 3.2$\times$ and 1.7$\times$ token reductions over Qwen3 backbones, significantly lowering on-device latency (Sec.~\ref{sec:tokeneff}). This efficiency stems from RL-side compression, primarily a soft overlong-response penalty that maintains accuracy while ensuring the training curriculum remains Pareto-dominant (App.~\ref{subsec:abl-tokenefficiency}).

\item \textbf{Quantized edge deployment.} At \texttt{Q4\_K\_M}, both model sizes retain aggregate score within one point of BF16 while reducing disk footprint by 68--69\%. On-device measurements connect \mqfourb{}'s shorter generations to 75\% lower median latency and estimated energy per answer at matched quantization, while peak RAM remains below 5~GB in the tested laptop and smartphone configurations. A broader six-format study identifies a \emph{capacity-vs-precision} effect at matched footprint (Sec.~\ref{sec:quant}; App.~\ref{app:quant}).
\end{itemize}

\section{Related Work}
\label{sec:related}

\textbf{Medical LLMs.} Med-PaLM~\cite{medpalm1} and Med-PaLM-2~\cite{medpalm} demonstrated that domain-specialized fine-tuning of large general LLMs could reach passing scores on USMLE-style exams, but they remain proprietary and large. Among open models, the strongest medical performance has largely come from substantially larger systems, such as MedGemma-27B~\cite{medgemma2025}, Meditron 70B-scale models~\cite{meditron}, and OpenBioLLM-70B~\cite{openbiollm}; smaller medical models such as MedGemma-1.5-4B-it~\cite{medgemma15} and JSL-MedLlama-3-8B~\cite{medllama} are more deployable but leave a clear accuracy gap. Based on our evaluation (App.~\ref{app:large-medical}) and to the best of our knowledge, no open model in the 1--4B range matched MedGemma-27B-text-it on closed-ended medical benchmarks or HealthBench prior to our work.

\textbf{Synthetic data for distillation.} Self-instruct and synthetic instruction generation~\cite{selfinstruct, evolinstruct} have become standard for scaling supervision. The Genesis III project~\cite{genesis3} focused on synthetic data for pre-training, combining seed acquisition, quality filtering, and a ``scaling Q\&A'' stage to produce broad educational corpora. We adapt this pipeline for medical post-training: selected Genesis III question/option fields and a new FineFineWeb-derived health slice are used as prompt seeds, and the resulting scaling-QA material is converted into SFT supervision by generating fresh medical reasoning targets with a stronger teacher model.

\textbf{RL for medical reasoning.} Fleming-R1~\cite{fleming-r1} applies a multi-stage GRPO curriculum with hard-sample mining to medical reasoning. AlphaMed~\cite{alphamedqa} releases the curated AlphaMed19K medical RL prompt set together with a binary correctness reward and a pass-at-$N$ difficulty annotation procedure. We adopt a two-stage curriculum in the spirit of Fleming-R1 and reuse AlphaMed's pass-at-$N$ annotation, applied here with each model's own size-matched SFT-2 policy as the annotator; we switch the optimiser to DAPO~\cite{dapo}, a GRPO variant that drops the KL anchor, uses asymmetric clip-higher, dynamically filters zero-variance rollout groups, and adds a soft overlong-response penalty.

\textbf{Reasoning-model test-time efficiency.} Reasoning models improve difficult tasks by generating explicit intermediate computation, but long traces increase decode latency, KV-cache memory~\cite{li2025thinkless}, and energy use~\cite{jin2025energy}, all especially costly on edge hardware. \medpsy{} targets this trade-off in the medical domain, improving accuracy while reducing generated tokens in compact reasoning models.

\textbf{Edge LLMs and quantization.} Edge deployment depends jointly on compressed model footprint and generation cost: quantization governs storage and memory requirements, while response length directly affects latency and energy consumption. GGUF and llama.cpp~\cite{llamacpp} support practical quantized inference across consumer hardware. Existing medical LLM studies primarily establish BF16 benchmark accuracy. We complement this literature by jointly evaluating quantized medical accuracy and measured on-device efficiency across two model sizes (Sec.~\ref{sec:quant}; App.~\ref{app:quant}).

\section{Data Methodology}
\label{sec:data}

In this section we describe at a high level the data used to post-train \medpsy{} and the teacher-selection process behind it. The training corpus description here reflects the design principles and the recipe used to train the final models that are evaluated in this paper.
\subsection{Training data overview}
\label{subsec:data-overview}
We explored several data mixtures and post-training methodologies before settling on the recipe used for the final \medpsy{} models. In total, \textbf{about 10M synthetic rows} of medical and healthcare supervision were generated for the final synthetic corpus. The recipe organizes the data into a \textbf{two-stage curriculum}: a broad-coverage corpus (Corpus 1) followed by a smaller, higher-value clinical-QA corpus (Corpus 2). The two corpora play complementary roles in the SFT curriculum (Sec.~\ref{sec:posttrain}).

Two principles guided the data construction throughout these ablations:
\begin{itemize}[leftmargin=*, itemsep=2pt, topsep=2pt]
\item \textbf{Synthetic datasets.} Corpus 1 utilizes broad synthetic prompts derived from two primary sources. First, we extracted the question and answer option fields from four public Genesis III subsets~\cite{genesis3}: \texttt{college\_medicine}, \texttt{professional\_medicine}, \texttt{college\_biology}, and \texttt{high\_school\_biology}. Second, following~\cite{genesis3}, we use Fleming-R1-32B~\cite{fleming-r1} to generate four multiple-choice questions (MCQs) for each document in the FineFineWeb health sub-domain~\cite{finefineweb}. We selected Fleming-R1-32B as it achieves strong performance on closed-ended medical tasks while offering higher throughput than larger teacher models. Corpus 2 instead utilizes train splits from focused open-source medical QA datasets as higher-value clinical prompts. Together, these broad synthetic and focused public-train prompts are used as seeds to generate the final synthetic training data.
\item \textbf{A single, controlled SFT target teacher.} Every long-form assistant target used for supervision (chain-of-thought traces, extended rationales, decision-oriented answers) is newly generated by Baichuan-M3-235B~\cite{baichuanm3}, the teacher selected in Sec.~\ref{subsec:teacher}, rather than taken from open-source assistant traces. This separation keeps prompt generation and target generation distinct: the seed data controls coverage, while Baichuan-M3-235B provides style adaptation and medical-reasoning knowledge distillation for the SFT targets.
\end{itemize}

The output of this stage is a curated medical and healthcare post-training corpus that combines synthetic expansion, reasoning-focused data, and public medical QA prompts, all paired with teacher-generated reasoning targets; source list, licenses, and prompts are in App.~\ref{app:sft-data}.

\subsection{Teacher selection}
\label{subsec:teacher}
Because our supervision \emph{is} the teacher's output, the teacher's ceiling directly bounds the student's. From a literature review and public medical leaderboards, we identified three strong candidates (\textbf{Baichuan-M3-235B}~\cite{baichuanm3}, \textbf{GPT-OSS-120B}~\cite{gptoss}, and \textbf{Fleming-R1-32B}~\cite{fleming-r1}) and evaluated them on our benchmark suite under three independent model graders (Sec.~\ref{subsec:eval-method}). The same suite serves teacher selection and final reporting, as it directly measures the supervision quality the student distills.

Baichuan-M3-235B leads with 74.83 average closed-ended accuracy (vs.\ 72.73/72.55), with its largest advantages on the most expert-level benchmarks (MedXpertQA: +5.61/+10.98; MMLU-Pro Health: +5.05/+2.73), and beats both alternatives by 6--12 HealthBench points across all three graders (Tables~\ref{tab:teacher-closed},~\ref{tab:teacher-hb}). Since the teacher's clinical reasoning and communication quality is exactly the signal the student distills, this open-ended advantage was decisive: we use Baichuan-M3-235B as the teacher for all final \medpsy{} training data. A controlled student-level ablation (App.~\ref{app:teacher}) verifies that this choice transfers when the prompt set, base model, and training pipeline are held fixed.

\section{Post-Training Methodology}
\label{sec:posttrain}

\subsection{Backbone selection}
We selected backbones via a controlled comparison at each target size against four other open candidates (full tables in App.~\ref{app:backbone}). At the 3-4B class, \textbf{Qwen3-4B-Thinking-2507} leads closed-ended by +13.43 over Llama-3.2-3B-Instruct, +14.11 over SmolLM3-3B, and +20.51 over gemma-3-4b-it, and dominates HealthBench by +21--26 points over Llama-3.2-3B-Instruct across all three graders. At the sub-2B class, \textbf{Qwen3-1.7B (Thinking)} leads by +5.81 over the next-best closed-ended backbone (LFM2.5-1.2B-Thinking), remains ahead of LFM2.5-1.2B-Thinking across all three HealthBench graders, and is +21--26 points ahead of Llama-3.2-1B-Instruct on HealthBench. We therefore build \mqfourb{} on top of Qwen3-4B-Thinking-2507 and \mqonesevenb{} on Qwen3-1.7B (with \texttt{enable\_thinking=True}); an architecture summary of both backbones is given in App.~\ref{app:archi}.

\subsection{Four-stage curriculum}
The four stages are:
\begin{enumerate}[leftmargin=*, itemsep=2pt, topsep=2pt]
\item \textbf{SFT Stage 1: broad medical adaptation} (Corpus 1). The model absorbs large-scale synthetic supervision spanning biology, medicine, and health, building wide factual coverage and medical vocabulary.
\item \textbf{SFT Stage 2: reasoning specialization} (Corpus 2). The model shifts to a smaller set of high-value clinical QA examples with teacher-generated CoT, sharpening answer structure and clinical reasoning depth. 
\item \textbf{RL Stage 1: consolidation on easy/moderate cases.} DAPO~\cite{dapo} on $\sim$16K examples from AlphaMed19K~\cite{alphamedqa}, initialized from the SFT-stage-2 checkpoint. This stage reinforces correct chains of reasoning on clinical scenarios where the model already has partial competence.
\item \textbf{RL Stage 2: hard-enriched targeting.} The full set is re-annotated with the best Stage-1 checkpoint and a hard-enriched subset of $\sim$2--4K items is built deterministically at a fixed 1:2 right-to-wrong ratio. The purpose of Stage 2 is to push the model on its persistent failure modes.
\end{enumerate}

Our RL phase builds on three lines of prior work. The two-stage RLVR-with-curriculum framework, consolidating reasoning on easier cases before targeting persistent failures, follows Fleming-R1~\cite{fleming-r1}. The Stage-1 corpus, the pass-at-$N$ difficulty annotation procedure, and the binary correctness reward are inspired by AlphaMed~\cite{alphamedqa}. The structured \texttt{<think>}/\texttt{<answer>} reasoning trace and the idea of conditioning the reward on the presence of a reasoning block follow Med-U1~\cite{medu1} (and DeepSeek-R1-Zero~\cite{deepseekr1} more broadly). We adopt the structured \texttt{<think>}/\texttt{<answer>} format and extend the binary correctness signal to a four-tier gated reward (\S\ref{para:reward}); a controlled ablation (App.~\ref{subsec:abl-effectiveness}) shows the four-tier extension is essentially neutral on accuracy versus a binary reward and provides only a modest token-efficiency benefit at 4B, so we report it as a design choice rather than a primary contribution. Within this scaffolding, our two RL design \emph{contributions} are: a single-pass deterministic Stage-2 mining procedure (\S\ref{para:mining}), and the integration of all four stages into a single training pipeline targeted at the sub-7B medical reasoning regime.

\paragraph{Difficulty annotation and Stage-1 data.}
Following AlphaMed, we annotate every training prompt empirically by sampling $N{=}5$ rollouts at temperature $0.7$ from the policy and labelling by the count $n_c$ of correct answers, mapped to six difficulty levels $L_1$--$L_6$ ($L_1{=}5/5$, $L_6{=}0/5$); the coarser Easy/Moderate/Difficult split groups $L_1$ alone, $L_2$--$L_5$ together, and $L_6$ alone. The Stage-1 training set follows AlphaMed's data recipe: all of MedQA plus per-level stratified sampling of MedMCQA (capped at 1{,}600 per $L_i$), with PubMedQA omitted following AlphaMed's finding that it harms reasoning transfer. From this, $\sim$16K items at the Easy/Moderate levels form the actual Stage-1 corpus.

\paragraph{Reasoning-aware reward.}
\label{para:reward}
The model is required to emit reasoning inside \texttt{<think>...</think>} tags and a final answer letter inside \texttt{<answer>...</answer>}~\cite{medu1, deepseekr1}. Let $c \in \{0,1\}$ indicate answer correctness (matching ground truth) and $t \in \{0,1\}$ indicate the presence of a non-trivial reasoning block. Our task reward generalises the binary correctness/format scheme of Med-U1 to a four-tier function that gates jointly on $c$ and $t$:
\begin{equation}
r_{\text{task}}(c, t) \;=\;
\begin{cases}
1.0 & \text{if } c=1, t=1 \quad (\text{correct }+\text{ reasoning})\\
0.5 & \text{if } c=1, t=0 \quad (\text{correct, no reasoning; ``shortcut''})\\
0.1 & \text{if } c=0, t=1 \quad (\text{reasoning, wrong answer; partial credit})\\
0.0 & \text{otherwise} \quad (\text{no valid structure}).
\end{cases}
\label{eq:reward-task}
\end{equation}
The 0.5 vs.\ 1.0 split actively penalises the shortcut of skipping reasoning even when the model happens to answer correctly; the 0.1 tier supplies a non-zero gradient on rollout groups where no candidate is correct, partially mitigating reward sparsity on Difficult items.

\paragraph{Single-pass deterministic Stage-2 mining.}
\label{para:mining}
Fleming-R1 adapts the Stage-2 distribution through iterative re-evaluation. We instead perform a single deterministic mining pass: after Stage 1, the best Stage-1 checkpoint is re-evaluated on the full 19K-item set with the same pass-at-$N$ procedure; items that the policy still fails on every attempt ($n_c{=}0$) are retained in full, and a smaller sample of items it gets right is added to enforce a 1:2 right-to-wrong ratio.\footnote{Concretely, our 4B Stage-2 set contains 2{,}439 items (4.7\% of the original set was still ``Difficult'' after Stage 1), and the 1.7B Stage-2 set contains 4{,}317 items (8.3\% Difficult after Stage 1).}

\paragraph{Optimization.}
Both RL stages use DAPO~\cite{dapo}, a GRPO variant whose actor objective drops the KL-to-reference term and replaces PPO's symmetric clip $\epsilon$ with an asymmetric $(\epsilon_{\text{low}}, \epsilon_{\text{high}})$:
\begin{equation}
\mathcal{L}_{\text{DAPO}}(\theta) \;=\; -\,\mathbb{E}_{(s,a)\sim\pi_{\theta_{\text{old}}}}\!\left[\,\min\!\Big(\rho_t(\theta)\,\hat{A}_t,\;\;\mathrm{clip}\!\left(\rho_t(\theta),\,1-\epsilon_{\text{low}},\,1+\epsilon_{\text{high}}\right)\hat{A}_t\Big)\right],
\label{eq:dapo}
\end{equation}
where $\rho_t(\theta)=\pi_\theta(a_t|s_t)/\pi_{\theta_{\text{old}}}(a_t|s_t)$ and $\hat{A}_t$ is the GRPO group-relative advantage computed from $n=16$ rollouts per prompt. We use $\epsilon_{\text{low}}=0.2$, $\epsilon_{\text{high}}=0.28$ (clip-higher) and dynamic filtering of zero-variance rollout groups (those for which all $n$ rollouts share the same reward, i.e.\ the advantage is identically zero). On top of $r_{\text{task}}$ in Eq.~\ref{eq:reward-task} we apply a soft overlong-response penalty over a $L_{\text{soft}}{=}2{,}048$-token buffer within an $L_{\text{max}}{=}4{,}096$-token response budget:
\begin{equation}
r_{\text{overlong}}(L) \;=\; -\,\lambda\,\max\!\left(0,\;\frac{L - L_{\text{soft}}}{L_{\text{max}} - L_{\text{soft}}}\right),
\qquad \lambda{=}1,
\label{eq:overlong}
\end{equation}
where $L$ is the per-rollout response length in tokens. The total per-rollout reward is $r = r_{\text{task}} + r_{\text{overlong}}$. Sequences that reach $L_{\text{max}}$ receive the full $-\lambda$, while sequences below $L_{\text{soft}}$ are unpenalised. Per-stage hyperparameters are in App.~\ref{app:compute}; Algorithm~\ref{alg:medpsy-pipeline} summarises the full pipeline.

\begin{algorithm}[t]
\caption{\medpsy{} four-stage post-training (one model size).}
\label{alg:medpsy-pipeline}
\begin{algorithmic}[1]
\Require backbone $B$; SFT corpora $\mathcal{D}_{\text{C1}}, \mathcal{D}_{\text{C2}}$; RL prompt set $\mathcal{D}_{\text{RL}}$ (AlphaMed19K)
\State $\theta_{\text{S1}} \gets \textsc{SFT}(B,\, \mathcal{D}_{\text{C1}})$ \Comment{SFT-1: broad medical adaptation}
\State $\theta_{\text{S2}} \gets \textsc{SFT}(\theta_{\text{S1}},\, \mathcal{D}_{\text{C2}})$ \Comment{SFT-2: reasoning specialization}
\State $(\mathcal{D}_{\text{em}},\, \mathcal{D}_{\text{hard}}) \gets \textsc{Annotate}(\theta_{\text{S2}},\, \mathcal{D}_{\text{RL}},\, N{=}5)$ \Comment{Pass@$N$ split: Easy/Mod vs.\ Hard}
\State $\theta_{\text{R1}} \gets \textsc{DAPO}\!\left(\theta_{\text{S2}},\, \mathcal{D}_{\text{em}};\; r_{\text{task}}{+}r_{\text{overlong}}\right)$ \Comment{RL Stage 1, Eq.~\ref{eq:reward-task}+\ref{eq:overlong}}
\State $\mathcal{D}_{\text{S2}} \gets \textsc{MineStage2}(\theta_{\text{R1}},\, \mathcal{D}_{\text{RL}})$ \Comment{Single-pass deterministic mining}
\State $\theta_{\text{R2}} \gets \textsc{DAPO}\!\left(\theta_{\text{R1}},\, \mathcal{D}_{\text{S2}};\; r_{\text{task}}{+}r_{\text{overlong}}\right)$ \Comment{RL Stage 2 (released \medpsy{} S2)}
\State \Return $\theta_{\text{R2}}$
\Statex
\Function{MineStage2}{$\pi,\, \mathcal{D}$}
  \For{$x \in \mathcal{D}$}\;
    $n_c(x) \gets \#\{\text{correct rollouts in } N{=}5 \text{ samples from } \pi \text{ at temperature } 0.7\}$
  \EndFor
  \State $\mathcal{D}_{\text{diff}} \gets \{x \in \mathcal{D} : n_c(x) = 0\}$ \Comment{Items $\pi$ still fails on every attempt}
  \State $\mathcal{D}_{\text{right}} \gets$ top-$\lfloor |\mathcal{D}_{\text{diff}}|/2 \rfloor$ items of $\{x : n_c(x) \ge 1\}$ ranked by $n_c(x)$ (descending)
  \State \Return $\mathcal{D}_{\text{diff}} \cup \mathcal{D}_{\text{right}}$ \Comment{$1{:}2$ right-to-wrong ratio}
\EndFunction
\end{algorithmic}
\end{algorithm}

This broad-then-focused design mirrors the SFT curriculum and is especially important in medicine, where strong average accuracy can mask weaknesses on rare or complex cases. Full training configurations/hyperparameters for SFT/RL and compute requirements are reported in Appendix~\ref{app:compute}. The recipe is not Qwen3-specific: applying all four stages to DeepSeek-R1-Distill-Qwen-1.5B~\cite{deepseekr1}, a Qwen2.5-derived reasoning checkpoint, lifts the seven-task closed-ended average from 28.4 to 52.2 and HealthBench from 7.0 to 58.0 (App.~\ref{app:backbone-transfer}).

\section{Evaluation}
\label{sec:eval}

\subsection{Methodology}
\label{subsec:eval-method}
\textbf{Generation protocol.} Unless otherwise noted, each model is evaluated with the generation parameters recommended for that model, and every reported evaluation is run three times with the arithmetic mean reported. For \medpsy{} and the Qwen3 baselines we use \texttt{temperature=0.6}, \texttt{top\_k=20}, \texttt{top\_p=0.95}, and \texttt{max\_output\_tokens=16384}.

\textbf{Closed-ended (LLM-as-Parser).} For all closed-ended benchmarks (MMLU-Health, MMLU-Pro Health, MedMCQA, MedQA-USMLE, MedXpertQA, AfriMedQA, PubMedQA), we use an LLM-as-Parser evaluation implemented in OpenCompass~\cite{opencompass}. MMLU-Health is reported as the averaged accuracy across six MMLU sub-domains: \texttt{anatomy}, \texttt{clinical\_knowledge}, \texttt{college\_biology}, \texttt{college\_medicine}, \texttt{medical\_genetics}, and \texttt{professional\_medicine}. The student model generates its full unconstrained response (with reasoning), and a separate reasoning model parses the final selected option, which is exact-matched against ground truth. This decouples generation from evaluation, avoiding both the suppression of reasoning that comes with letter-only constrained decoding and the false negatives that plague regex-based extraction. Main closed-ended tables use QwQ~\cite{qwq} as the parser; an example parser prompt is in App.~\ref{app:sft-data}. To estimate parser sensitivity, we re-scored the same generations with three independent parsers (Llama-3.3-70B-Instruct~\cite{llama33}, GPT-OSS-120B~\cite{gptoss}, and QwQ); the maximum parser-induced average-score spread is only 0.45 points across all evaluated models, with stable model rankings and coherent per-benchmark differences (App.~\ref{app:eval-robustness}).

\textbf{HealthBench (physician-authored rubrics with model-based grading).} HealthBench~\cite{healthbench} evaluates 5{,}000 realistic health conversations using 48{,}562 conversation-specific criteria written by 262 physicians. Each criterion specifies what a response should include or avoid and carries a physician-assigned importance weight; model graders apply this fixed expert-authored standard to each response. The benchmark's model-grading protocol was validated against physician judgments. We evaluate with three independent grader families (\textbf{CompassJudger-2-32B-Instruct}~\cite{compassjudger}, \textbf{Llama-3.3-70B-Instruct}~\cite{llama33}, and \textbf{GPT-OSS-120B}~\cite{gptoss}) to test whether conclusions depend on a particular grader. Table~\ref{tab:hb-judges} reports all three directly, while per-dimension breakdowns use CompassJudger-2-32B-Instruct (App.~\ref{app:hb-dimensions}). Three-grader results for teacher and backbone selection are reported in App.~\ref{app:teacher}--\ref{app:backbone}. HealthBench-Hard is the hardest subset, requiring multi-step reasoning, complex safety judgments, and expert-level clinical knowledge.

\textbf{Train--test decontamination.} To account for overlaps in public medical datasets, we reran the closed-ended evaluation after excluding matched benchmark test items. Decontamination removed $157$ examples ($1.10\%$) without altering model rankings (App.~\ref{app:eval-robustness}).
\subsection{Closed-ended results}
\label{subsec:closed}

\begin{table}[t]
\caption{\textbf{Closed-ended benchmark results}, sorted by average. Average is over seven closed-ended benchmarks; bold marks best per column within each size class.}
\label{tab:closed}
\centering
\small
\setlength{\tabcolsep}{2pt}
\adjustbox{max width=\linewidth}{%
\begin{tabular}{@{}lcccccccc@{}}
\toprule
\textbf{Model} & \textbf{Avg} & \makecell{\textbf{MMLU}\\ \textbf{Health}} & \textbf{AfriMedQA} & \makecell{\textbf{MMLU-Pro}\\ \textbf{Health}} & \textbf{MedMCQA} & \makecell{\textbf{MedQA}\\ \textbf{(USMLE)}} & \textbf{MedXpertQA} & \textbf{PubMedQA} \\
\midrule
\bestcell{\mqfourb{}} & \bestcell{70.54} & 89.70 & 71.50 & 70.45 & 72.15 & \bestcell{84.39} & \bestcell{30.61} & \bestcell{75.00} \\
MedGemma-27B-text-it & 69.95 & \bestcell{90.48} & \bestcell{73.07} & \bestcell{72.94} & \bestcell{72.77} & 83.29 & 25.18 & 71.93 \\
Qwen3-4B-Thinking-2507 & 63.10 & 85.92 & 64.12 & 67.73 & 61.78 & 70.91 & 16.69 & 74.53 \\
\midrule
\bestcell{\mqonesevenb{}} & \bestcell{62.62} & \bestcell{82.72} & \bestcell{64.84} & \bestcell{61.37} & \bestcell{63.56} & \bestcell{75.05} & \bestcell{21.28} & 69.53 \\
MedGemma-1.5-4B-it & 51.20 & 67.69 & 54.38 & 47.31 & 50.08 & 64.39 & 15.80 & 58.73 \\
Qwen3-1.7B (Thinking) & 49.95 & 72.49 & 51.87 & 45.07 & 49.14 & 47.18 & 11.60 & \bestcell{72.33} \\
\bottomrule
\end{tabular}}
\end{table}

\mqfourb{} matches MedGemma-27B-text-it in aggregate closed-ended performance (70.54 vs.\ 69.95; +0.59) at 6.75$\times$ smaller scale. A matched-instance analysis following Medmarks v0.1.0~\cite{medmarks} likewise finds near parity across three runs (50.25\% versus 49.75\%; App.~\ref{app:pairwise-winrate}). Compared to its own backbone, post-training adds \textbf{+7.44} average. The lead is not explained by parser choice or train--test overlap: the ranking is stable across three closed-ended parsers and remains after decontamination (App.~\ref{app:eval-robustness}). \mqonesevenb{} reaches 62.62, beating MedGemma-1.5-4B-it by +11.42 and matching Qwen3-4B-Thinking-2507 at less than half its size; the largest gains over MedGemma-1.5-4B are MMLU-Health (+15.03) and MedQA-USMLE (+10.66). PubMedQA is a known ceiling case: the Qwen3-1.7B backbone already scores 72.33, so the 1.7B post-training trades $-2.80$ on PubMedQA for double-digit gains elsewhere; the 4B model avoids this regression and improves PubMedQA to 75.00. Accuracy also carries over to much longer inputs: on LongHealth~\cite{adams2024longhealth} (5{,}090--6{,}754-word clinical documents), \mqfourb{} scores 0.855/0.875 on Tasks 1/2, above MedGemma-27B-text-it (0.828/0.802), and \mqonesevenb{} leads MedGemma-1.5-4B-it by up to 30.6 points (App.~\ref{app:longhealth}).

\subsection{HealthBench results}
\label{subsec:healthbench}
HealthBench complements closed-ended accuracy by evaluating responses to realistic, open-ended clinical scenarios against fixed, conversation-specific physician-authored rubrics. Its seven dimensions cover expertise-tailored communication, response depth, context seeking, emergency referrals, global health, health-data tasks, and responding under uncertainty. Table~\ref{tab:hb-judges} reports Standard and Hard scores under all three grader families.

\begin{table}[t]
\caption{\textbf{HealthBench overall scores under three independent model graders, Standard (Std) and Hard splits.} Bold marks the best score per column. Per-dimension breakdowns under CompassJudger-2-32B-Instruct are in App.~\ref{app:hb-dimensions} (Tab.~\ref{tab:hb-std-dim}--\ref{tab:hb-hard-dim}).}
\label{tab:hb-judges}
\centering
\small
\setlength{\tabcolsep}{4pt}
\adjustbox{max width=\linewidth}{%
\begin{tabular}{lcccccc}
\toprule
& \multicolumn{2}{c}{\textbf{CompassJudger-2-32B}} & \multicolumn{2}{c}{\textbf{Llama-3.3-70B}} & \multicolumn{2}{c}{\textbf{GPT-OSS-120B}} \\
\cmidrule(lr){2-3} \cmidrule(lr){4-5} \cmidrule(lr){6-7}
\textbf{Model} & \textbf{Std} & \textbf{Hard} & \textbf{Std} & \textbf{Hard} & \textbf{Std} & \textbf{Hard} \\
\midrule
\bestcell{\mqfourb{}}      & \bestcell{74.00} & \bestcell{58.00} & \bestcell{66.33} & \bestcell{48.33} & \bestcell{51.33} & \bestcell{28.67} \\
\mqonesevenb{}             & 70.33            & 54.33            & 63.00            & 46.00            & 46.00            & 24.67 \\
MedGemma-27B-text-it       & 65.00            & 42.67            & 59.00            & 36.00            & 44.67            & 13.00 \\
Qwen3-4B-Thinking-2507     & 63.00            & 42.00            & 56.00            & 34.00            & 36.67            & \phantom{0}9.33 \\
MedGemma-1.5-4B-it         & 54.00            & 29.67            & 48.00            & 24.67            & 31.00            & \phantom{0}2.00 \\
\bottomrule
\end{tabular}}
\end{table}

\mqfourb{} ranks first on every (grader, split) cell of Tab.~\ref{tab:hb-judges}, and \mqonesevenb{} is second-best across the board, surpassing MedGemma-27B-text-it on both Standard and Hard under all three graders. Absolute scores vary with grader strictness, but the complete five-model ordering is identical across all six grader-by-split cells, and the Hard-split lead is uniformly larger than the Standard-split lead under every grader. The gains therefore do not depend on the calibration of a single model grader (cross-grader consistency summary in App.~\ref{app:judges}).

A per-dimension breakdown under CompassJudger-2-32B-Instruct is reported in App.~\ref{app:hb-dimensions}: \mqfourb{} leads on all seven dimensions in both Standard and Hard evaluations, while \mqonesevenb{} surpasses MedGemma-27B-text-it on all seven HealthBench-Hard dimensions.

\subsection{Stage-wise progression}
\label{subsec:stages}
\begin{table}[t]
\caption{\textbf{Training stage progression.} Comparison of \mqonesevenb{} (1.7B) and \mqfourb{} (4B) across all benchmarks. Avg is the closed-ended mean; HB is HealthBench Standard (CompassJudger-2-32B).}
\label{tab:stages}
\centering
\footnotesize
\setlength{\tabcolsep}{1.0pt}
\adjustbox{max width=\linewidth}{%
\begin{tabular}{@{}l cc cc cc cc cc cc cc cc cc@{}}
\toprule
& \multicolumn{2}{c}{\textbf{Avg}} & \multicolumn{2}{c}{\textbf{HB}} & \multicolumn{2}{c}{\makecell{\textbf{MMLU}\\\textbf{Health}}} & \multicolumn{2}{c}{\makecell{\textbf{Afri}\\\textbf{MedQA}}} & \multicolumn{2}{c}{\makecell{\textbf{MMLU-Pro}\\\textbf{Health}}} & \multicolumn{2}{c}{\textbf{MedMCQA}} & \multicolumn{2}{c}{\textbf{MedQA}} & \multicolumn{2}{c}{\makecell{\textbf{MedXpert}\\\textbf{QA}}} & \multicolumn{2}{c}{\makecell{\textbf{PubMed}\\\textbf{QA}}} \\
\cmidrule(lr){2-3} \cmidrule(lr){4-5} \cmidrule(lr){6-7} \cmidrule(lr){8-9} \cmidrule(lr){10-11} \cmidrule(lr){12-13} \cmidrule(lr){14-15} \cmidrule(lr){16-17} \cmidrule(lr){18-19}
\textbf{Stage} & \textbf{1.7B} & \textbf{4B} & \textbf{1.7B} & \textbf{4B} & \textbf{1.7B} & \textbf{4B} & \textbf{1.7B} & \textbf{4B} & \textbf{1.7B} & \textbf{4B} & \textbf{1.7B} & \textbf{4B} & \textbf{1.7B} & \textbf{4B} & \textbf{1.7B} & \textbf{4B} & \textbf{1.7B} & \textbf{4B} \\
\midrule
Backbone & 49.95 & 63.10 & 53.00 & 63.00 & 72.49 & 85.92 & 51.87 & 64.12 & 45.07 & 67.73 & 49.14 & 61.78 & 47.18 & 70.91 & 11.60 & 16.69 & 72.33 & 74.53 \\
+ SFT-1  & 57.48 & 67.93 & 70.33 & 74.00 & 79.70 & 88.77 & 60.90 & 69.35 & 57.05 & 68.87 & 57.26 & 67.72 & 63.97 & 82.38 & 17.09 & 26.22 & 66.40 & 72.20 \\
+ SFT-2  & 59.70 & 69.29 & 70.33 & 74.00 & 80.14 & 89.58 & 62.45 & 70.47 & 60.27 & 69.77 & 58.54 & 69.21 & 68.76 & 84.32 & 18.89 & 28.71 & 68.87 & 73.00 \\
+ RL-1   & 60.00 & 68.85 & 70.33 & 74.00 & 80.92 & 89.32 & 63.90 & 71.23 & 59.66 & 69.56 & 61.33 & 71.21 & 68.21 & 84.13 & 16.26 & 25.47 & 69.73 & 71.07 \\
+ RL-2   & \bestcell{62.62} & \bestcell{70.54} & \bestcell{70.33} & \bestcell{74.00} & \bestcell{82.72} & \bestcell{89.70} & \bestcell{64.84} & \bestcell{71.50} & \bestcell{61.37} & \bestcell{70.45} & \bestcell{63.56} & \bestcell{72.15} & \bestcell{75.05} & \bestcell{84.39} & \bestcell{21.28} & \bestcell{30.61} & 69.53 & \bestcell{75.00} \\
\midrule
\textbf{Total $\Delta$} & \textbf{+12.67} & \textbf{+7.44} & \textbf{+17.33} & \textbf{+11.00} & +10.23 & +3.78 & +12.97 & +7.38 & +16.30 & +2.72 & +14.42 & +10.37 & +27.87 & +13.48 & +9.68 & +13.92 & $-$2.80 & +0.47 \\
\bottomrule
\end{tabular}}
\end{table}
Several patterns are consistent across both sizes (Table~\ref{tab:stages}). \emph{SFT Stage 1 delivers the largest single jump} (+7.53 / +4.83), confirming that broad medical exposure from Corpus 1 is the foundation; gains are most dramatic on HealthBench (+17.33 / +11.00) and MedQA-USMLE (+16.79 / +11.47). \emph{SFT Stage 2 adds focused reasoning} on MedQA-USMLE (+4.79 for 1.7B), MMLU-Pro Health (+3.22 for 1.7B), and MedXpertQA (+2.49 for 4B). \emph{The two RL stages together provide the final sharpening}, peaking on MedQA-USMLE (+6.29 for 1.7B) and MedMCQA (+5.02 / +2.94), the benchmarks that most reward structured clinical reasoning. {Stage~2 hard-mining accounts for essentially all of the released model's MedXpertQA lift: replacing the mined Stage-2 set with a uniformly-random subset of identical size loses $5.0$~pp (4B) and $2.0$~pp (1.7B) on MedXpertQA, while easy/medium benchmarks are essentially unchanged (App.~\ref{subsec:abl-effectiveness}).} \emph{HealthBench gains are installed early}: SFT Stage 1 already lifts HealthBench to its final level, and the RL stages preserve it while improving closed-ended accuracy and token efficiency.

\subsection{Token efficiency}
\label{sec:tokeneff}

Sample-weighted across all eight benchmarks, the 4B model achieves a \textbf{3.2$\times$} reduction in average response length (\textbf{909} vs.\ \textbf{2{,}953} tokens for Qwen3-4B-Thinking-2507) while \emph{improving} accuracy, and the 1.7B model a \textbf{1.7$\times$} reduction (\textbf{1{,}110} vs.\ \textbf{1{,}901} tokens for Qwen3-1.7B Thinking). Complementing these observed length reductions, an explicit thinking-budget sweep shows that \mqfourb{} retains 96.6\% of its unbounded closed-ended score with a 300-token thinking cap (App.~\ref{app:thinking-budget}). The 1.7B reduction is smaller because its backbone already emits shorter outputs, and on HealthBench \mqonesevenb{} writes \emph{slightly longer} responses than its backbone, trading tokens for the +17 HealthBench points of Sec.~\ref{subsec:stages}. The per-benchmark breakdown is in App.~\ref{app:tokens-detail} (Tab.~\ref{tab:tokens-detail}).

We decompose the RL-side compression into three independently controllable mechanisms in App.~\ref{subsec:abl-tokenefficiency}: vanilla GRPO with a KL anchor compresses by only $\sim 6$--$20$\%, switching to DAPO's drift-enabling modifications (KL-removal and asymmetric clip-higher) brings compression to $\sim 44$--$45$\%, and adding the soft overlong-response penalty pushes it to the released-model $\sim 65$--$66$\%. The \emph{ordering} of the RL stages compounds this: Stage~1 trains on items typically answerable with short reasoning, installing a low-token-budget habit; Stage~2 then pushes accuracy on hard items starting from this already-compressed checkpoint and preserves the short-reasoning habit.

\subsection{Quantization for mobile deployment}
\label{sec:quant}
To meet mobile RAM constraints, we convert both models to GGUF with llama.cpp~\cite{llamacpp} and evaluate six quantization formats spanning 8-, 4-, and 3-bit precision. This sweep identifies \texttt{Q4\_K\_M} as the preferred deployment configuration, offering the strongest observed balance between footprint reduction and full-suite score retention at both model sizes. Table~\ref{tab:q4-deployment} reports the complete per-format comparison together with the real-device profile at \texttt{Q4\_K\_M}; the imatrix ablation and per-benchmark details are in App.~\ref{app:quant}.

\begin{table}[t]
\begingroup
\color{black}
\caption{\textbf{Quantized deployment profile.} \textbf{(a)} Complete quantized GGUF variants comparison. $\Delta$ Size relative to BF16; $\Delta$ Score is the absolute change in $(\text{HB Overall}+\text{CE Avg})/2$. \textbf{(b)} Apple M5 Pro measurements at matched \texttt{Q4\_K\_M} quantization over 30 prompts (20 MedQA and 10 MedXpertQA-Text); time per answer is the median, and energy per answer is estimated as sustained SoC power (26.5 W) times median wall time.}
\label{tab:q4-deployment}
\label{tab:quant-variants}
\centering
\small
\setlength{\tabcolsep}{4pt}
\textbf{(a) Complete six-format comparison}\\[2pt]
\adjustbox{max width=\linewidth}{%
\begin{tabular}{lc cc cc cc cc cc cc}
\toprule
 & & \multicolumn{2}{c}{\textbf{Size (GB)}} & \multicolumn{2}{c}{\textbf{$\Delta$ Size}} & \multicolumn{2}{c}{\textbf{HB}} & \multicolumn{2}{c}{\textbf{HB-H}} & \multicolumn{2}{c}{\textbf{CE Avg}} & \multicolumn{2}{c}{\textbf{$\Delta$ Score}} \\
\cmidrule(lr){3-4} \cmidrule(lr){5-6} \cmidrule(lr){7-8} \cmidrule(lr){9-10} \cmidrule(lr){11-12} \cmidrule(lr){13-14}
\textbf{Variant} & \textbf{imat} & \textbf{4B} & \textbf{1.7B} & \textbf{4B} & \textbf{1.7B} & \textbf{4B} & \textbf{1.7B} & \textbf{4B} & \textbf{1.7B} & \textbf{4B} & \textbf{1.7B} & \textbf{4B} & \textbf{1.7B} \\
\midrule
BF16 & --- & 8.3 & 3.8 & --- & --- & 74 & 70 & 58 & 54 & 70.54 & 62.62 & --- & --- \\
Q8\_0 & no & 4.4 & 2.1 & $-$47\% & $-$45\% & 74 & 70 & 57 & 55 & 70.25 & 62.62 & $-$0.15 & 0.00 \\
Q4\_K\_M & yes & 2.6 & 1.2 & $-$69\% & $-$68\% & 73 & 69 & 56 & 52 & 69.92 & 62.16 & $-$0.81 & $-$0.73 \\
IQ4\_NL & yes & 2.5 & 1.2 & $-$70\% & $-$68\% & 73 & 69 & 57 & 51 & 69.50 & 60.22 & $-$1.02 & $-$1.70 \\
IQ4\_XS & yes & 2.4 & 1.1 & $-$71\% & $-$71\% & 73 & 69 & 57 & 53 & 69.39 & 60.05 & $-$1.08 & $-$1.79 \\
IQ3\_M & yes & 2.0 & 0.98 & $-$76\% & $-$74\% & 73 & 67 & 58 & 49 & 68.55 & 58.46 & $-$1.50 & $-$3.58 \\
IQ3\_XXS & yes & 1.8 & 0.85 & $-$78\% & $-$78\% & 69 & 59 & 51 & 40 & 64.42 & 48.71 & $-$5.56 & $-$12.46 \\
\bottomrule
\end{tabular}}\\[6pt]
\textbf{(b) Matched \texttt{Q4\_K\_M} local inference}\\[2pt]
\adjustbox{max width=\linewidth}{%
\begin{tabular}{lccccc}
\toprule
\textbf{Model} & \textbf{Peak RAM} & \textbf{Decode} & \textbf{Tokens/answer} & \textbf{Time/answer} & \textbf{Energy/answer} \\
\midrule
\mqfourb{}                  & 4.29 GB & 83.9 tok/s & 738   & 8.75 s & 232 J \\
Qwen3-4B-Thinking-2507      & 4.28 GB & 82.2 tok/s & 2{,}727 & 34.6 s & 917 J \\
\bottomrule
\end{tabular}}\par
\endgroup
\end{table}

\noindent

\noindent\textbf{\texttt{Q4\_K\_M} accuracy and footprint.} At \texttt{Q4\_K\_M}, both model sizes reduce disk footprint by 68--69\% while keeping aggregate score within one point of BF16 (Table~\ref{tab:q4-deployment}(a)). \mqfourb{} reaches 2.6~GB with a 0.81-point aggregate-score change, while \mqonesevenb{} reaches 1.2~GB with a 0.73-point change. Across the broader six-format study, the 4B model retains strong accuracy through 3-bit quantization, and the matched-footprint comparison reveals a capacity-vs-precision effect: the 2.0~GB \texttt{IQ3\_M} 4B model outperforms the 2.1~GB \texttt{Q8\_0} 1.7B model (Table~\ref{tab:q4-deployment}(a)). At matched on-disk size, allocating storage to more parameters at lower precision buys more performance than fewer parameters at higher precision.

\noindent\textbf{Full quantization landscape.} Beyond \texttt{Q4\_K\_M}, the six-format comparison (Table~\ref{tab:q4-deployment}(a); Fig.~\ref{fig:quant-sizes}) shows that \texttt{Q8\_0} is near-lossless at both sizes ($-$0.15 and 0.00 aggregate-score change at roughly half the footprint). The I-quants trade precision for footprint gracefully at 4B: \texttt{IQ4\_NL} and \texttt{IQ4\_XS} lose 1.0--1.1 points, and \texttt{IQ3\_M} (2.0~GB) loses only 1.50 while exactly matching the BF16 HealthBench-Hard score of 58. Collapse occurs only at \texttt{IQ3\_XXS} ($-$5.56 at 4B, $-$12.46 at 1.7B), and even this worst configuration (64.42 closed-ended, 69 HealthBench) remains above the unquantized Qwen3-4B-Thinking-2507 backbone (63.10/63) and MedGemma-1.5-4B-it (51.20/54). The 1.7B model degrades roughly twice as fast as the 4B below 4-bit precision, consistent with the intuition that smaller models have less weight redundancy: every channel carries more of the model's behavior, so precision lost per channel translates more directly into degraded reasoning. We therefore do not recommend any sub-4-bit variant of \mqonesevenb{} for medical use.

\noindent\textbf{Real-device deployment.} At matched \texttt{Q4\_K\_M} quantization, \mqfourb{} and its Qwen3 backbone have comparable per-token decode rates, while \mqfourb{} generates 73\% fewer tokens per answer. This translates to 74.7\% lower median end-to-end latency and estimated energy per answer on the Apple M5 Pro, with 4.29~GB peak RAM (Table~\ref{tab:q4-deployment}(b)). A separate 8K-context stress test records 4.83~GB as the largest laptop peak for \mqfourb{} \texttt{Q4\_K\_M} (App.~\ref{app:context-cost}). Measurements on a OnePlus 13 confirm fully local execution at 2.22~GB peak RAM for \mqonesevenb{} and 3.98~GB for \mqfourb{}; full protocols and measurements are in App.~\ref{app:device}.
\section{Discussion and Limitations}
\label{sec:limitations}

The results support our central thesis: for compact medical LLMs, medical capability and deployment efficiency can be optimized jointly. In \medpsy{}, broad SFT establishes medical coverage, focused SFT improves clinical reasoning, and hard-sample RL sharpens the most reasoning-heavy benchmarks (Sec.~\ref{subsec:stages}). The full curriculum (SFT$\to$SFT$\to$RL Stage~1$\to$RL Stage~2) is also strictly Pareto-dominant over single-stage hard-only RL on token-vs-accuracy (best performance at $\sim 3$--$4\times$ shorter response length) (App.~\ref{subsec:abl-tokenefficiency}). The Easy/Moderate Stage~1 leaves the policy already compressed and high-reward before Stage~2 mining targets residual hard items, preserving response compression even in the more challenging scenarios that require longer reasoning.

However, these findings should be interpreted within several important boundaries: the models are \textbf{text-only}; the training data is predominantly English. Like all LLMs, the models can produce confident but incorrect medical statements; they should not be used for autonomous diagnosis or treatment. Real clinical decision support would require additional safety review, IRB-grade evaluation, prospective testing, and regulatory clearance.

\section{Conclusion and Future Work}
\label{sec:conclusion}
These limitations point to several directions for future work. The recipe should be extended to multilingual and multimodal medical settings, and evaluation should move beyond the current benchmark suite. In particular, compact medical models need stronger open-ended evaluations, safety benchmarks, and medical error detection tests before their behavior can be understood in higher-risk clinical scenarios.

Within this scope, \medpsy{} shows that compact, text-only medical LLMs (1.7B and 4B) can match or surpass much larger medical models: \mqfourb{} matches MedGemma-27B on closed-ended benchmarks at 6.75$\times$ smaller scale, and both sizes surpass it on HealthBench-Hard. Both run fully locally within 5~GB of peak RAM on the tested laptop and smartphone, where shorter generations cut median latency and estimated energy per answer by 75\%. A controlled synthetic-data pipeline, a broad-then-focused SFT curriculum, and hard-sample RL deliver strong medical accuracy and improved token efficiency ($1.7$--$3.2\times$ fewer generated tokens) at small scale. The quantization study further shows that \texttt{Q4\_K\_M} keeps aggregate score within one point of BF16 at a 68--69\% smaller footprint, while very low precision exposes a clear capacity-vs-precision trade-off. Overall, \medpsy{} suggests that for compact medical LLMs, \emph{data quality, staged post-training, and alignment design dominate parameter scale, and medical capability and deployment efficiency can be optimized jointly.}

\begin{ack}
\begingroup
\renewcommand{\thefootnote}{*}
This research was supported and funded by Tether AI Research\footnote{References to Tether AI Research are references to Tether Data, S.A. de C.V.} which provided computational resources and financial compensation for the authors. No other party provided funding to support this research.

All authors are contractors engaged by or for Tether AI Research. The authors declare no other competing financial or non-financial interests.
\endgroup
\end{ack}

\small
\bibliographystyle{unsrtnat}
\bibliography{bibliography}
\normalsize

\clearpage
\appendix
\section{Technical appendices and supplementary material}

\let\oldsubsection\subsection
\renewcommand{\subsection}{\FloatBarrier\oldsubsection}

\begingroup
\etocsettocdepth{subsection}
\renewcommand{\contentsname}{Appendix contents}
\localtableofcontents
\endgroup

\subsection{Data sources and prompts}
\label{app:sft-data}
Corpus~1 uses selected Genesis III biology/medicine seeds plus new health seeds. For the health slice, we select FineFineWeb health-domain passages after Ultra-FineWeb-classifier\cite{ultrafineweb} filtering and apply the scaling-QA prompt below to generate self-contained MCQ user prompts. We keep only the question/options as the user turn, then use Baichuan-M3-235B to generate every assistant target; Corpus~2 uses only the public train sources listed in Tab.~\ref{tab:sft-data-mix} as the user turn, then uses Baichuan-M3-235B as well to generate every assistant target.

\begin{table}[!htbp]
\caption{\textbf{Final Data Mix and Source Licenses.} SFT row counts represent final selected input for assistant target generation; RL counts describe the AlphaMed19K prompt pool used for DAPO.}
\label{tab:sft-data-mix}
\centering
\footnotesize
\renewcommand{\arraystretch}{1.2}
\begin{tabular}{@{} l p{5cm} r l @{}}
\toprule
\textbf{Group} & \textbf{Source / Role} & \textbf{Rows} & \textbf{License / Note} \\
\midrule

\multicolumn{4}{l}{\textit{Corpus 1: Broad SFT \& Seeds}} \\
Seeds & \quad Genesis III (Biology/Medicine) & 8,089,891 & CC-BY-NC-4.0 \\
Seeds & \quad FineFineWeb (Health)        & 2,133,642 & Apache-2.0 \\
SFT-1 & \quad Synthetic Broad Med Corpus  & 10,223,533 & Inherited \\
\addlinespace[0.5em]

\multicolumn{4}{l}{\textit{Corpus 2: Focused SFT}} \\
SFT-2 & \quad AfriMedQA MCQ Train         & 124 & CC-BY-4.0 \\
SFT-2 & \quad MedMCQA Train               & 129,231 & Apache-2.0 \\
SFT-2 & \quad MedQA-USMLE Train           & 9,176 & CC-BY-4.0 \\
\addlinespace[0.5em]

\multicolumn{4}{l}{\textit{RL Prompt Pool}} \\
RL    & \quad AlphaMed19K    & $\sim$19,000 & MIT \\
\bottomrule
\end{tabular}
\end{table}

\paragraph{Scaling-QA prompt for new health-domain seeds.}
The prompt template used for generating samples starting from the seeds is as follows:
\begin{lstlisting}[basicstyle=\scriptsize\ttfamily,frame=single,breaklines=true,breakatwhitespace=true,columns=fullflexible,keepspaces=true,aboveskip=10pt]
You are a {{level}} {{domain}} tutor. I will give you a context passage.
IGNORE the details of the passage. Use it only as inspiration for topics, but DO NOT refer to the passage, book, text, or "discussed/mentioned" material in your questions.
Context:
{{output}}

Generate EXACTLY 4 independent, self-contained questions suitable for {{level}} level {{domain}}.

Output rules (MUST be followed strictly):
- Output EXACTLY 4 samples.
- Each row = ONE question in the following format:
  "<question>","<choiceA>","<choiceB>","<choiceC>","<choiceD>","<correct_choice_letter>"
- Each row MUST end with a newline (line break).
- Do NOT join multiple questions into one line.
- Do NOT output any extra text, headers, or blank lines.

Field definitions:
- <question> = a fully self-contained {{level}}-level {{domain}} problem.
  It must NOT reference any passage, book, figure, table, or "as discussed" or "as mentioned" style wording.
  Include all necessary values, constants, or definitions directly inside the question.
- <choiceA> ... <choiceD> = four mutually exclusive, plausible answer choices.
- <correct_choice_letter> = exactly one of A, B, C, D.

Question variety:
- Use a mix of imperative (e.g., "Calculate..."), interrogative, completion/fill-in, and true/false style.
- Do NOT start all questions with the same word.
- Include realistic values/constants when needed (e.g., g = 9.8 m/s^2).
- Use standard SI units and symbols.

Hard constraints:
- Generate exactly 4 questions (4 samples).
- Correct answers must be evenly distributed: one A, one B, one C, one D.
- Randomize placement of correct answers; no clustering.
- Each row must contain exactly 6 comma-separated values.
- Do NOT add quotes around the entire block; only around individual fields.

Output ONLY the 4 samples in the required format, nothing else.
\end{lstlisting}

\paragraph{Closed-ended answer-extraction prompt.}
This is the MMLU extractor; the same template was used for each closed-ended dataset, changing only the option list and dataset-specific field names.

\begin{lstlisting}[basicstyle=\scriptsize\ttfamily,frame=single,breaklines=true,breakatwhitespace=true,columns=fullflexible,keepspaces=true,aboveskip=10pt]
MMLU_ANSWER_EXTRACTOR_TEMPLATE = """
You are an expert answer extractor. Your ONLY job is to extract the final answer from the candidate's response.

CRITICAL INSTRUCTIONS - READ CAREFULLY:
- DO NOT solve the question yourself
- DO NOT generate a new answer
- DO NOT think about what the correct answer should be
- DO NOT evaluate whether the candidate's answer is right or wrong
- ONLY extract what the candidate actually wrote as their final answer

Your task is purely extraction, not generation or evaluation.

Here are the extraction guidelines for MMLU multiple choice questions:
1. Look for the candidate's final answer in their response. This should be one of: A, B, C, or D
   - Look for explicit statements like "ANSWER: A", "The answer is B", "Final answer: C"
   - Look for \boxed{A} format (extract what's inside the braces)
   - Look for standalone letters A, B, C, or D in their conclusion
   - The final choice they settle on in their reasoning

2. If the candidate's response contains multiple different answers:
   - If they state one answer but then correct/fix themselves, extract the corrected/final answer they provided
   - If they state multiple different answers without choosing one or correcting themselves, return "MULTIPLE_ANSWERS"

3. If you cannot clearly identify a single letter choice (A, B, C, or D) in the response, return "NO_ANSWER"

4. If the candidate generates new questions:
   - DO NOT extract the answer from the new generated questions
   - ONLY extract the answer from the original question

RESPONSE FORMAT:
First, provide a brief explanation of why you are extracting that particular answer (what indicators you found in the candidate's response).
Then, provide the extracted answer as a single letter.

Use this exact format:

Extraction Reasoning: [Brief explanation of what indicators led you to extract this answer from the candidate's response]

Extracted Candidate's Answer: [A single letter: A, B, C, or D. Use MULTIPLE_ANSWERS if the candidate provided multiple different answers, or NO_ANSWER if no clear answer was found.]

<Original Question Begin>:
Question: {input}

A. {A}
B. {B}
C. {C}
D. {D}

Answer:
<Original Question End>

<Candidate's Response Begin>:
{prediction}
<Candidate's Response End>
"""
\end{lstlisting}

\subsection{Teacher-selection ablation}
\label{app:teacher}
This appendix details the ablation study performed to determine the most effective teacher model for our distillation pipeline. We evaluate candidate models across several closed-ended medical benchmarks in Table \ref{tab:teacher-closed} and report their overall HealthBench performance across three distinct graders in Table \ref{tab:teacher-hb}.

As shown in the results, Baichuan-M3-235B consistently outperforms the other candidates, which justifies its selection as the primary teacher for our experiments.

\begin{table}[!htbp]
\caption{Teacher closed-ended benchmark results. Bold marks best per column.}
\label{tab:teacher-closed}
\centering
\footnotesize
\setlength{\tabcolsep}{3pt}
\adjustbox{max width=\linewidth}{%
\begin{tabular}{lcccccccc}
\toprule
Teacher & Avg & AfriMedQA & MMLU-H & MedQA & PubMedQA & MedMCQA & MedXpertQA & MMLU-Pro-H \\
\midrule
\textbf{Baichuan-M3-235B} & \textbf{74.83} & \textbf{74.58} & \textbf{93.08} & 88.90 & 73.27 & \textbf{76.88} & \textbf{40.91} & \textbf{76.20} \\
GPT-OSS-120B & 72.73 & 73.28 & 91.85 & \textbf{89.03} & 75.20 & 73.26 & 35.30 & 71.15 \\
Fleming-R1-32B & 72.55 & 72.68 & 92.61 & 85.91 & \textbf{79.20} & 74.02 & 29.93 & 73.47 \\
\bottomrule
\end{tabular}}
\end{table}

\begin{table}[!htbp]
\caption{Teacher HealthBench overall scores across three graders.}
\label{tab:teacher-hb}
\centering
\footnotesize
\begin{tabular}{lccc}
\toprule
Teacher & CompassJudger & Llama-3.3-70B & GPT-OSS-120B \\
\midrule
\textbf{Baichuan-M3-235B} & \textbf{77} & \textbf{71} & \textbf{58} \\
GPT-OSS-120B & 69.67 & 63 & 52.33 \\
Fleming-R1-32B & 67 & 61 & 45.67 \\
\bottomrule
\end{tabular}
\end{table}

\paragraph{Student-level validation of teacher choice.} To test whether the teacher-level HealthBench advantage transfers to the student, we trained three controlled 4B SFT arms from the same base model on the same 170K-prompt subset with the same training pipeline, changing only the source of the target responses. Table~\ref{tab:teacher-student} reports the results.

\begin{table}[!htbp]
\caption{\textbf{Student-level teacher ablation} (4B students, identical 170K prompts, varying only the SFT target source). Closed-ended averages differ by at most 0.41 points, but HealthBench separates the arms by 11 points; the mixed model stays near Fleming-only rather than at the 65.25 midpoint of the single-teacher arms.}
\label{tab:teacher-student}
\centering
\footnotesize
\begin{tabular}{lcc}
\toprule
\textbf{SFT target source} & \textbf{Closed-ended avg} & \textbf{HealthBench} \\
\midrule
Fleming-R1-32B only & 63.43 & 59.5 \\
Baichuan-M3-235B only & 63.02 & \textbf{71.0} \\
Both teachers, mixed & 63.40 & 60.0 \\
\bottomrule
\end{tabular}
\end{table}

Closed-ended averages differ by at most 0.41 points, whereas HealthBench separates the arms by 11 points. Mixing heterogeneous target teachers provides no benefit here: the mixed arm (60.0) tracks the Fleming-only arm (59.5) rather than the midpoint of the two single-teacher arms (65.25), consistent with a less coherent SFT signal. In this controlled run, the Baichuan-only arm therefore best preserves the open-ended criterion used for teacher selection.

\subsection{Additional large medical baselines}
\label{app:large-medical}

Table~\ref{tab:large-medical-baselines} reports the closed-ended benchmark comparison against additional open medical LLMs in the 8--70B range.

\begin{table}[!htbp]
\caption{\textbf{Closed-ended comparison with additional open medical baselines.} Average is computed over the seven closed-ended benchmarks.}
\label{tab:large-medical-baselines}
\centering
\footnotesize
\setlength{\tabcolsep}{3pt}
\adjustbox{max width=\linewidth}{%
\begin{tabular}{lcccccccc}
\toprule
Model & Avg & AfriMedQA & MMLU-H & MedQA & PubMedQA & MedMCQA & MedXpertQA & MMLU-Pro-H \\
\midrule
\textbf{\mqfourb{}} & \textbf{70.54} & 71.50 & 89.70 & \textbf{84.39} & 75.00 & 72.15 & \textbf{30.61} & 70.45 \\
MedGemma-27B-text-it & 69.95 & \textbf{73.07} & \textbf{90.48} & 83.29 & 71.93 & 72.77 & 25.18 & \textbf{72.94} \\
Meditron3-70B & 69.57 & 71.65 & 88.93 & 80.65 & \textbf{80.07} & 70.74 & 23.66 & 71.31 \\
OpenBioLLM-70B & 67.46 & 67.46 & 90.10 & 76.12 & 77.93 & \textbf{72.92} & 19.22 & 68.46 \\
Meditron3-8B & 57.51 & 60.21 & 76.55 & 62.74 & 74.47 & 58.35 & 14.39 & 55.87 \\
JSL-MedLlama-3-8B-v2.0 & 55.03 & 54.12 & 77.92 & 56.35 & 74.47 & 57.97 & 11.92 & 52.45 \\
MedGemma-1.5-4B-it & 51.20 & 54.38 & 67.69 & 64.39 & 58.73 & 50.08 & 15.80 & 47.31 \\
\bottomrule
\end{tabular}}
\end{table}

\subsection{Backbone-selection ablation}
\label{app:backbone}

We evaluated multiple candidate backbones at each target size on our full medical benchmark suite \emph{before} any post-training. Tables~\ref{tab:bb-4b-closed},~\ref{tab:bb-4b-hb},~\ref{tab:bb-1b-closed},~\ref{tab:bb-1b-hb} report the results.

\begin{table}[!htbp]
\caption{4B-class backbone closed-ended results.}
\label{tab:bb-4b-closed}
\centering
\footnotesize
\setlength{\tabcolsep}{3pt}
\adjustbox{max width=\linewidth}{%
\begin{tabular}{lcccccccc}
\toprule
Model & Avg & MMLU-H & AfriMedQA & MMLU-Pro-H & MedMCQA & MedQA & MedXpertQA & PubMedQA \\
\midrule
\textbf{Qwen3-4B-Thinking-2507} & \textbf{63.10} & \textbf{85.92} & \textbf{64.12} & \textbf{67.73} & \textbf{61.78} & \textbf{70.91} & \textbf{16.69} & 74.53 \\
Qwen3-4B (Thinking) & 60.64 & 83.89 & 63.15 & 61.82 & 60.21 & 67.64 & 15.01 & 72.80 \\
Llama-3.2-3B-Instruct & 49.67 & 66.81 & 52.39 & 39.97 & 52.35 & 49.51 & 12.43 & \textbf{74.20} \\
SmolLM3-3B & 48.99 & 72.64 & 47.24 & 47.23 & 46.20 & 46.40 & 11.32 & 71.93 \\
gemma-3-4b-it & 42.59 & 62.72 & 45.46 & 34.68 & 45.08 & 40.51 & 10.52 & 59.13 \\
\bottomrule
\end{tabular}}
\end{table}

\begin{table}[!htbp]
\caption{4B-class backbone HealthBench results across three graders.}
\label{tab:bb-4b-hb}
\centering
\footnotesize
\begin{tabular}{lccc}
\toprule
Model & CompassJudger & Llama-3.3-70B & GPT-OSS-120B \\
\midrule
\textbf{Qwen3-4B-Thinking-2507} & \textbf{63} & \textbf{56} & 36.67 \\
Qwen3-4B (Thinking) & 62 & 55 & \textbf{37.67} \\
gemma-3-4b-it & 59 & 53 & 33.33 \\
SmolLM3-3B & 50 & 43.67 & 24 \\
Llama-3.2-3B-Instruct & 37.33 & 32 & 14.67 \\
\bottomrule
\end{tabular}
\end{table}

\begin{table}[!htbp]
\caption{Sub-2B-class backbone closed-ended results.}
\label{tab:bb-1b-closed}
\centering
\footnotesize
\setlength{\tabcolsep}{3pt}
\adjustbox{max width=\linewidth}{%
\begin{tabular}{lcccccccc}
\toprule
Model & Avg & MMLU-H & AfriMedQA & MMLU-Pro-H & MedMCQA & MedQA & MedXpertQA & PubMedQA \\
\midrule
\textbf{Qwen3-1.7B (Thinking)} & \textbf{49.95} & \textbf{72.49} & \textbf{51.87} & \textbf{45.07} & \textbf{49.14} & \textbf{47.18} & \textbf{11.60} & \textbf{72.33} \\
LFM2.5-1.2B-Thinking & 44.15 & 63.48 & 45.07 & 37.81 & 42.11 & 39.85 & 11.54 & 69.20 \\
Llama-3.2-1B-Instruct & 36.18 & 49.34 & 36.84 & 23.88 & 37.89 & 34.04 & 10.25 & 61.00 \\
SmolLM2-1.7B-Instruct & 33.32 & 49.27 & 31.39 & 20.70 & 33.80 & 29.22 & 9.86 & 59.00 \\
\bottomrule
\end{tabular}}
\end{table}

\begin{table}[!htbp]
\caption{Sub-2B-class backbone HealthBench results across three graders.}
\label{tab:bb-1b-hb}
\centering
\footnotesize
\begin{tabular}{lccc}
\toprule
Model & CompassJudger & Llama-3.3-70B & GPT-OSS-120B \\
\midrule
\textbf{Qwen3-1.7B (Thinking)} & \textbf{53} & \textbf{47.33} & \textbf{27.67} \\
LFM2.5-1.2B-Thinking & 49 & 41.67 & 22.33 \\
Llama-3.2-1B-Instruct & 25 & 22 & 5.33 \\
SmolLM2-1.7B-Instruct & 23 & 18.67 & 6 \\
\bottomrule
\end{tabular}
\end{table}

Among the tested sub-2B candidates, Qwen3-1.7B provided the strongest pretrained starting point across the closed-ended suite and all three HealthBench graders, motivating its selection for the final model.

\subsection{Transfer to a Qwen2.5-derived reasoning checkpoint}
\label{app:backbone-transfer}
To test the portability of our post-training recipe beyond Qwen3, we applied the same four-stage pipeline to DeepSeek-R1-Distill-Qwen-1.5B~\cite{deepseekr1}, a compact reasoning model built on Qwen2.5-Math and distilled from DeepSeek-R1. Table~\ref{tab:deepseek-transfer} shows consistent gains at every training stage on both the seven closed-ended benchmarks and HealthBench, demonstrating that the recipe also transfers effectively to a Qwen2.5-derived reasoning checkpoint.

\begin{table}[!htbp]
\caption{Stage-wise transfer results for DeepSeek-R1-Distill-Qwen-1.5B. Scores have weighted mean output tokens from the same protocol in parentheses. The explicit-reasoning rows add a ``Think step by step'' instruction and required final-answer format; the paper-mirror rows do not.}
\label{tab:deepseek-transfer}
\centering
\footnotesize
\adjustbox{max width=\linewidth}{%
\begin{tabular}{lccccc}
\toprule
Metric & Base & SFT-1 & SFT-2 & RL-1 & RL-2 \\
\midrule
Closed-ended avg.\ (7), paper mirror & 28.42 (1,209) & 44.07 (2,430) & 46.96 (2,394) & 49.25 (1,107) & 52.16 (1,601) \\
Closed-ended avg.\ (7), explicit reasoning & 29.53 (3,007) & 45.76 (2,730) & 48.20 (2,664) & 49.92 (1,068) & 53.60 (1,476) \\
HealthBench & 7.0 (1,292) & 55.0 (2,530) & 56.0 (2,450) & 57.0 (2,071) & 58.0 (2,089) \\
\bottomrule
\end{tabular}}
\end{table}

\paragraph{Evaluation behavior.} During evaluation, we observed that DeepSeek-R1-Distill-Qwen-1.5B behaves differently depending on whether the prompt explicitly requests step-by-step reasoning. We therefore report two closed-ended settings: the paper-mirror setting uses the same prompts and answer format as the main experiments, whereas the explicit-reasoning setting adds ``Think step by step'' and requires the specified final-answer format. The explicit-reasoning setting is applied only to the closed-ended benchmarks; HealthBench is always evaluated under the standard protocol, without any added reasoning instruction or required answer format.

The recipe improves accuracy consistently in both settings. Under the paper-mirror evaluation, the seven-task average rises from 28.42 to 52.16, with gains on all seven tasks; after RL, output length is 33.1\% below SFT-2 (1,601 vs.\ 2,394 tokens). Under explicit reasoning, the average rises from 29.53 to 53.60 while output length falls from 3,007 to 1,476 tokens, a 50.9\% reduction. The two settings thus reach comparable final accuracy while exhibiting different length behavior. Under the standard HealthBench protocol, the score rises from 7.0 to 58.0, with output length increasing from 1,292 to 2,089 tokens. Overall, the transfer experiment shows strong gains across both elicitation styles, while explicit prompting primarily changes the token-efficiency profile.

\subsection{Backbone architectures}
\label{app:archi}

Table~\ref{tab:arch} summarizes the architectural parameters of the two selected Qwen3 backbones.

\begin{table}[!htbp]
\caption{Architecture summary of the chosen Qwen3 backbones.}
\label{tab:arch}
\centering
\footnotesize
\begin{tabular}{lcc}
\toprule
Parameter & Qwen3-1.7B & Qwen3-4B \\
\midrule
Hidden size & 2{,}048 & 2{,}560 \\
FFN hidden size & 6{,}144 & 9{,}728 \\
Layers & 28 & 36 \\
Attention heads & 16 & 32 \\
KV groups (GQA) & 8 & 8 \\
Vocab size & 151{,}936 & 151{,}936 \\
Position embedding & RoPE & RoPE \\
Normalization & RMSNorm & RMSNorm \\
Activation & SwiGLU & SwiGLU \\
\bottomrule
\end{tabular}
\end{table}

\subsection{Compute resources}
\label{app:compute}
\textbf{Cluster.} 30 worker nodes; 8$\times$NVIDIA H100~80GB HBM3 per node; InfiniBand interconnect (Mellanox ConnectX, 8 HCA ports/node); SLURM scheduler. Containers: NVIDIA NeMo~25.09 (Enroot), PyTorch~2.5.1, CUDA~12.1.

\textbf{Per-job parallelism.} 1.7B SFT: 4 nodes, 32 GPUs, TP$=$1, PP$=$1, DP$=$32. 4B SFT: 4 nodes, 32 GPUs, TP$=$1, PP$=$1, DP$=$32.

\textbf{Training hyperparameters.} Both SFT runs used 3 epochs, sequence length 4{,}096 with packed sequence size 4{,}096, global batch size 512 packed sequences, and micro batch size 1. Optimization used distributed fused Adam in bf16 mixed precision with a learning rate of $2\times10^{-5}$, minimum learning rate of $2\times10^{-6}$, cosine decay, and a linear warmup over 1\% of training steps. For RL training, effective per-update batch is $512$ prompts $\times$ $n{=}16$ rollouts $=8{,}192$ sampled trajectories. We present the RL configuration in Table~\ref{tab:rl-hparams}.

\begin{table}[!htbp]
    \caption{\textbf{RL training hyperparameters (DAPO).}}
    \label{tab:rl-hparams}
    \centering
    \footnotesize
    \setlength{\tabcolsep}{3pt}
    \begin{tabular}{l c c c c}
    \toprule
    Hyperparameter & S1 (1.7B) & S1 (4B) & S2 (1.7B) & S2 (4B) \\
    \midrule
    Asymmetric clip $(\epsilon_{\text{low}},\epsilon_{\text{high}})$ & \multicolumn{4}{c}{$(0.20,\;0.28)$} \\
    Entropy coefficient & \multicolumn{4}{c}{$0$} \\
    Optimizer / precision & \multicolumn{4}{c}{AdamW, bf16 mixed} \\
    Learning rate & \multicolumn{4}{c}{$1\!\times\!10^{-6}$ (constant)} \\
    Train batch (prompts) & \multicolumn{4}{c}{$512$} \\
    PPO mini-batch & \multicolumn{4}{c}{$128$} \\
    PPO micro-batch / GPU & \multicolumn{4}{c}{$8$} \\
    Gradient checkpointing & \multicolumn{4}{c}{enabled} \\
    Rollouts per prompt $n$ & \multicolumn{4}{c}{$16$} \\
    Max prompt / response length & \multicolumn{4}{c}{$512$ / $4{,}096$ tokens} \\
    vLLM max\_model\_len & \multicolumn{4}{c}{$4{,}608$ tokens} \\
    Tensor parallel (vLLM) & $1$ & $2$ & $1$ & $2$ \\
    Buffer length $L_{\text{soft}}$ / cap $L_{\text{max}}$ & \multicolumn{4}{c}{$2{,}048$ / $4{,}096$ tokens} \\
    Penalty factor $\lambda$ & \multicolumn{4}{c}{$1.0$} \\
    \bottomrule
    \end{tabular}
    \end{table}

\textbf{Tracking and checkpoints.} Weights \& Biases plus TensorBoard. Checkpoints in Megatron \texttt{torch\_dist} format (fully parallel save), converted to HuggingFace \texttt{safetensors} for evaluation and deployment.

\textbf{Training time.} The primary experimental pipeline required approximately 8,250 H100 GPU hours. The vast majority of this budget was allocated to the data generation phase ($\sim$8,000 hours), while downstream training stages were significantly less compute-intensive; Supervised Fine-Tuning (SFT) and Reinforcement Learning (RL) consumed approximately 100 and 150 hours, respectively. These figures represent the final end-to-end training of our primary model. Including preliminary ablations, failed runs, and evaluation, the total computational expenditure for the project is estimated at approximately 30,000 H100 GPU hours.

\subsection{Per-\texorpdfstring{grader}{grader} HealthBench results}
\label{app:judges}
HealthBench's evaluation criteria are written by physicians; the model graders in our panel determine whether each response satisfies those fixed, conversation-specific criteria. The main paper reports overall scores under all three graders (CompassJudger-2-32B-Instruct, Llama-3.3-70B-Instruct, GPT-OSS-120B) directly in Tab.~\ref{tab:hb-judges}. Three-grader results are also reported for the teacher- and backbone-selection ablations (App.~\ref{app:teacher},~\ref{app:backbone}). \\
Across all four uses of the panel (final-model comparison, teacher ablation, 4B-class backbone ablation, sub-2B-class backbone ablation), the main conclusions are preserved across the three graders: the selected teacher and backbone remain at or near the top, with only a small GPT-OSS swap between the two Qwen3-4B variants, and the released \medpsy{} models keep the same ordering. Absolute-score levels are uniformly highest under CompassJudger-2-32B-Instruct, intermediate under Llama-3.3-70B-Instruct, and lowest under GPT-OSS-120B (the strictest of the three graders on open-ended clinical content).

\subsection{HealthBench per-dimension and per-axis breakdowns}
\label{app:hb-dimensions}
Tab.~\ref{tab:hb-std-dim} and Tab.~\ref{tab:hb-hard-dim} report the per-dimension breakdown of HealthBench (Standard and Hard splits) under CompassJudger-2-32B-Instruct, the strongest dedicated model grader on evaluator benchmarks. The seven dimensions are expertise-tailored communication, response depth, context seeking, emergency referrals, global health, health-data tasks, and responding under uncertainty. The Overall column matches the corresponding CompassJudger column of Tab.~\ref{tab:hb-judges}. The 4B model's strongest dimensions are Emergency Referrals (81.67/62.33), Expertise-Tailored Communication (79.33/55.33), and Responding under Uncertainty (76.33/61.00; Standard/Hard).

\begin{table}[!htbp]
\caption{\textbf{HealthBench (Standard) per dimension}, CompassJudger-2-32B-Instruct.}
\label{tab:hb-std-dim}
\centering
\footnotesize
\setlength{\tabcolsep}{3pt}
\adjustbox{max width=\linewidth}{%
\begin{tabular}{lcccccccc}
\toprule
\textbf{Model} & \textbf{Overall} & \makecell{\textbf{Expert.}\\\textbf{Comm.}} & \makecell{\textbf{Resp.}\\\textbf{Depth}} & \makecell{\textbf{Context}\\\textbf{Seeking}} & \makecell{\textbf{Emerg.}\\\textbf{Refer.}} & \makecell{\textbf{Global}\\\textbf{Health}} & \makecell{\textbf{Health}\\\textbf{Data}} & \makecell{\textbf{Resp.\ under}\\\textbf{Uncert.}} \\
\midrule
\textbf{\mqfourb{}} & \textbf{74.00} & \textbf{79.33} & \textbf{63.67} & \textbf{71.67} & \textbf{81.67} & \textbf{73.67} & \textbf{60.67} & \textbf{76.33} \\
\mqonesevenb{} & 70.33 & 76.33 & 56.33 & 69.33 & 80.00 & 68.33 & 57.00 & 74.00 \\
MedGemma-27B-text-it & 65.00 & 73.00 & 61.33 & 58.67 & 73.00 & 61.00 & 56.67 & 66.33 \\
Qwen3-4B-Thinking-2507 & 63.00 & 71.00 & 58.00 & 57.67 & 74.00 & 59.00 & 54.67 & 64.33 \\
MedGemma-1.5-4B-it & 54.00 & 62.67 & 48.67 & 46.00 & 64.00 & 47.67 & 44.67 & 58.33 \\
Qwen3-1.7B (Thinking) & 53.00 & 63.67 & 49.67 & 48.33 & 64.67 & 45.67 & 42.33 & 56.33 \\
\bottomrule
\end{tabular}}
\end{table}

\begin{table}[!htbp]
\caption{\textbf{HealthBench-Hard per dimension}, CompassJudger-2-32B-Instruct.}
\label{tab:hb-hard-dim}
\centering
\footnotesize
\setlength{\tabcolsep}{3pt}
\adjustbox{max width=\linewidth}{%
\begin{tabular}{lcccccccc}
\toprule
\textbf{Model} & \textbf{Overall} & \makecell{\textbf{Expert.}\\\textbf{Comm.}} & \makecell{\textbf{Resp.}\\\textbf{Depth}} & \makecell{\textbf{Context}\\\textbf{Seeking}} & \makecell{\textbf{Emerg.}\\\textbf{Refer.}} & \makecell{\textbf{Global}\\\textbf{Health}} & \makecell{\textbf{Health}\\\textbf{Data}} & \makecell{\textbf{Resp.\ under}\\\textbf{Uncert.}} \\
\midrule
\textbf{\mqfourb{}} & \textbf{58.00} & \textbf{55.33} & \textbf{47.67} & \textbf{63.33} & \textbf{62.33} & \textbf{60.00} & \textbf{46.67} & \textbf{61.00} \\
\mqonesevenb{} & 54.33 & 52.33 & 40.33 & 61.00 & 60.33 & 55.00 & 43.33 & 58.33 \\
MedGemma-27B-text-it & 42.67 & 44.67 & 38.67 & 42.00 & 39.67 & 42.67 & 39.33 & 42.67 \\
Qwen3-4B-Thinking-2507 & 42.00 & 45.00 & 38.67 & 43.00 & 47.33 & 43.33 & 39.67 & 42.00 \\
MedGemma-1.5-4B-it & 29.67 & 31.67 & 29.00 & 28.00 & 29.00 & 29.00 & 23.67 & 35.00 \\
Qwen3-1.7B (Thinking) & 28.33 & 31.67 & 28.33 & 32.00 & 27.67 & 26.67 & 21.33 & 31.00 \\
\bottomrule
\end{tabular}}
\end{table}

\paragraph{Per-axis breakdown.} Where the per-dimension tables show \emph{where} failures occur, Table~\ref{tab:hb-failure-axes} shows \emph{how} response quality deteriorates: it reports the five behavior axes defined by the native HealthBench evaluator. Values are percentages on the 1,000-example Hard split, scored with CompassJudger-2-32B-Instruct and averaged over three independent generation runs. Context Awareness and Instruction Following are the two lowest-scoring axes for both \medpsy{} model sizes.

\begin{table}[!htbp]
\caption{\textbf{HealthBench-Hard behavior-axis scores} under CompassJudger-2-32B-Instruct. Values are three-run means over the 1,000-example Hard split.}
\label{tab:hb-failure-axes}
\centering
\footnotesize
\adjustbox{max width=\linewidth}{%
\begin{tabular}{lccccc}
\toprule
Model & Accuracy & Completeness & Context awareness & Instruction following & Communication quality \\
\midrule
\mqfourb{} & 63.5 & 64.1 & 50.7 & 53.4 & 71.3 \\
\mqonesevenb{} & 59.8 & 61.0 & 48.0 & 45.0 & 69.2 \\
MedGemma-27B-text-it & 57.2 & 43.9 & 23.0 & 54.6 & 71.4 \\
Qwen3-4B-Thinking-2507 & 54.8 & 44.4 & 25.4 & 58.5 & 74.6 \\
MedGemma-1.5-4B-it & 45.5 & 29.6 & 11.8 & 46.5 & 71.7 \\
Qwen3-1.7B (Thinking) & 42.5 & 31.0 & 9.5 & 50.1 & 66.7 \\
\bottomrule
\end{tabular}}
\end{table}

\subsection{Per-benchmark response length}
\label{app:tokens-detail}

Table~\ref{tab:tokens-detail} reports the per-benchmark average output length. The first row reports the weighted average; the remaining rows are the per-benchmark values, sorted by Qwen3-4B-Thinking-2507 response length (descending). Reductions are \mqfourb{} / \mqonesevenb{} versus the corresponding Qwen3 backbone; values $< 1{\times}$ indicate that \medpsy{} produces \emph{longer} responses than the backbone on that benchmark. The only such case is HealthBench at the 1.7B class ($0.9{\times}$), reflecting a deliberate trade for the +17~pp HealthBench gain reported in Sec.~\ref{subsec:stages}.

\begin{table}[!htbp]
\caption{\textbf{Per-benchmark average response length} (tokens; lower is better).}
\label{tab:tokens-detail}
\centering
\footnotesize
\setlength{\tabcolsep}{3pt}
\adjustbox{max width=\linewidth}{%
\begin{tabular}{l ccc c ccc}
\toprule
& \multicolumn{3}{c}{\textbf{4B class}} & & \multicolumn{3}{c}{\textbf{1.7B class}} \\
\cmidrule(lr){2-4} \cmidrule(lr){6-8}
\textbf{Benchmark} & \makecell{Qwen3-4B-\\Thinking-2507} & \mqfourb{} & \textbf{Reduction} & & \makecell{Qwen3-1.7B\\(Thinking)} & \mqonesevenb{} & \textbf{Reduction} \\
\midrule
\textbf{Weighted avg.\ (all)} & \textbf{2{,}953} & \bestcell{\textbf{909}}   & \bestcell{\textbf{3.2$\times$}} & & \textbf{1{,}901} & \bestcell{\textbf{1{,}110}} & \bestcell{\textbf{1.7$\times$}} \\
\midrule
MedXpertQA       & 4{,}900 & \bestcell{1{,}152} & \bestcell{4.3$\times$} & & 2{,}641 & \bestcell{1{,}459} & \bestcell{1.8$\times$} \\
MedQA-USMLE      & 4{,}050 & \bestcell{810}     & \bestcell{5.0$\times$} & & 2{,}886 & \bestcell{1{,}128} & \bestcell{2.6$\times$} \\
MMLU-Pro Health  & 3{,}396 & \bestcell{773}     & \bestcell{4.4$\times$} & & 2{,}564 & \bestcell{994}     & \bestcell{2.6$\times$} \\
AfriMedQA        & 2{,}969 & \bestcell{660}     & \bestcell{4.5$\times$} & & 2{,}079 & \bestcell{875}     & \bestcell{2.4$\times$} \\
PubMedQA         & 2{,}703 & \bestcell{527}     & \bestcell{5.1$\times$} & & 998     & \bestcell{788}     & \bestcell{1.3$\times$} \\
MMLU (Health)    & 2{,}534 & \bestcell{581}     & \bestcell{4.4$\times$} & & 2{,}020 & \bestcell{749}     & \bestcell{2.7$\times$} \\
MedMCQA          & 2{,}387 & \bestcell{597}     & \bestcell{4.0$\times$} & & 1{,}636 & \bestcell{770}     & \bestcell{2.1$\times$} \\
HealthBench      & 2{,}222 & \bestcell{1{,}405} & \bestcell{1.6$\times$} & & \bestcell{1{,}327} & 1{,}534 & 0.9$\times$ \\
\bottomrule
\end{tabular}}
\end{table}

\subsection{Evaluation Robustness and Decontamination}

\label{app:eval-robustness}

This appendix provides two robustness checks for the primary closed-ended results. First, we assess parser reliability by re-parsing the original model generations using three independent LLM parsers (Tab.~\ref{tab:parser-robustness}). Second, we address potential train--test leakage by rerunning the evaluation suite after excluding test examples that overlap with our final synthetic training corpus (Tabs.~\ref{tab:decontam-results} and~\ref{tab:decontam-deltas}).

\paragraph{Decontamination Methodology and Impact.} Although our pipeline utilizes only public training splits, medical datasets are frequently not mutually disjoint. Our audit revealed cross-dataset overlaps: for example, MedQA training items appeared in the MedXpertQA test set, and MedMCQA training items overlapped with AfriMedQA test items. To ensure the integrity of our results, we performed a string-matching decontamination and excluded all overlapping test examples from the final scoring.

The decontaminated results utilize the same checkpoints as the original evaluation; only the overlapping benchmark examples are removed. As shown in the following tables, the percentage of excluded examples is minimal. The impact on performance across all models was found to be negligible, and the relative rankings between our models and the baselines remain unchanged. Full per-benchmark exclusion counts and the resulting decontaminated scores are detailed in Tables~\ref{tab:decontam-results} and~\ref{tab:decontam-deltas}.

\begin{table}[!htbp]
\caption{\textbf{Closed-ended parser robustness.} Main-paper closed-ended tables use QwQ as the parser; this ablation re-scores the same generations with Llama-3.3-70B-Instruct, GPT-OSS-120B, and QwQ. Average is over the full seven-benchmark closed-ended suite.}
\label{tab:parser-robustness}
\centering
\footnotesize
\setlength{\tabcolsep}{3pt}
\adjustbox{max width=\linewidth}{%
\begin{tabular}{llcccccccc}
\toprule
\textbf{Model} & \textbf{Parser} & \textbf{Avg} & \textbf{AfriMedQA} & \textbf{MMLU-H} & \textbf{MedQA} & \textbf{PubMedQA} & \textbf{MedMCQA} & \textbf{MedXpertQA} & \textbf{MMLU-Pro-H} \\
\midrule
MedGemma-27B-text-it & Llama & 70.33 & 73.34 & 90.80 & 84.29 & 72.20 & 72.92 & 25.31 & 73.45 \\
MedGemma-27B-text-it & GPT-OSS & 70.22 & 73.31 & 90.64 & 84.11 & 72.20 & 72.84 & 25.31 & 73.13 \\
MedGemma-27B-text-it & QwQ & 69.95 & 73.07 & 90.48 & 83.29 & 71.93 & 72.77 & 25.18 & 72.94 \\
MedGemma-27B-text-it & MAX--MIN & 0.38 & 0.27 & 0.32 & 1.00 & 0.27 & 0.15 & 0.13 & 0.51 \\
\midrule
Qwen3-4B-Thinking-2507 & Llama & 63.49 & 64.32 & 86.18 & 71.95 & 74.73 & 62.06 & 17.07 & 68.12 \\
Qwen3-4B-Thinking-2507 & GPT-OSS & 63.27 & 64.26 & 86.10 & 71.20 & 74.73 & 61.92 & 16.72 & 67.96 \\
Qwen3-4B-Thinking-2507 & QwQ & 63.10 & 64.12 & 85.92 & 70.91 & 74.53 & 61.78 & 16.69 & 67.73 \\
Qwen3-4B-Thinking-2507 & MAX--MIN & 0.39 & 0.20 & 0.26 & 1.04 & 0.20 & 0.28 & 0.38 & 0.39 \\
\midrule
MedGemma-1.5-4B-it & Llama & 51.65 & 54.55 & 67.92 & 65.23 & 59.47 & 50.59 & 15.86 & 47.93 \\
MedGemma-1.5-4B-it & GPT-OSS & 51.20 & 53.95 & 67.65 & 64.99 & 58.73 & 49.95 & 15.84 & 47.29 \\
MedGemma-1.5-4B-it & QwQ & 51.20 & 54.38 & 67.69 & 64.39 & 58.73 & 50.08 & 15.80 & 47.31 \\
MedGemma-1.5-4B-it & MAX--MIN & 0.45 & 0.60 & 0.27 & 0.84 & 0.74 & 0.64 & 0.06 & 0.64 \\
\midrule
Qwen3-1.7B (Thinking) & Llama & 50.09 & 51.99 & 72.60 & 47.47 & 72.27 & 49.25 & 11.65 & 45.40 \\
Qwen3-1.7B (Thinking) & GPT-OSS & 50.07 & 51.93 & 72.62 & 47.47 & 72.47 & 49.20 & 11.63 & 45.17 \\
Qwen3-1.7B (Thinking) & QwQ & 49.95 & 51.87 & 72.49 & 47.18 & 72.33 & 49.14 & 11.60 & 45.07 \\
Qwen3-1.7B (Thinking) & MAX--MIN & 0.13 & 0.12 & 0.13 & 0.29 & 0.20 & 0.11 & 0.05 & 0.33 \\
\midrule
\mqfourb{} & Llama & 70.89 & 71.71 & 89.78 & 85.47 & 75.10 & 72.55 & 30.84 & 70.78 \\
\mqfourb{} & GPT-OSS & 70.85 & 71.70 & 90.07 & 85.15 & 75.07 & 72.37 & 30.74 & 70.85 \\
\mqfourb{} & QwQ & 70.54 & 71.50 & 89.70 & 84.39 & 75.00 & 72.15 & 30.61 & 70.45 \\
\mqfourb{} & MAX--MIN & 0.35 & 0.21 & 0.37 & 1.08 & 0.10 & 0.40 & 0.23 & 0.40 \\
\midrule
\mqonesevenb{} & Llama & 62.94 & 65.12 & 83.04 & 75.44 & 69.73 & 63.93 & 21.37 & 61.95 \\
\mqonesevenb{} & GPT-OSS & 62.89 & 65.02 & 83.04 & 75.44 & 69.73 & 63.84 & 21.39 & 61.77 \\
\mqonesevenb{} & QwQ & 62.62 & 64.84 & 82.72 & 75.05 & 69.53 & 63.56 & 21.28 & 61.37 \\
\mqonesevenb{} & MAX--MIN & 0.32 & 0.28 & 0.32 & 0.39 & 0.20 & 0.37 & 0.11 & 0.58 \\
\bottomrule
\end{tabular}}
\end{table}

\paragraph{Matched-instance pairwise comparison.}
\label{app:pairwise-winrate}
Following the original Medmarks v0.1.0 implementation~\cite{medmarks}, we assign one point to the model with the higher reward on each matched example and half a point to each model for a tie, average these scores within each dataset, and combine them using log-dataset-size weighting. The analysis covers AfriMedQA, MMLU-Health, MMLU-Pro Health, MedMCQA, MedQA-USMLE, MedXpertQA, and PubMedQA, with the six MMLU-Health subsets pooled before weighting. The three-run result in Table~\ref{tab:pairwise-winrate} confirms matching closed-ended performance.

\begin{table}[!htbp]
\caption{Matched-instance pairwise scores by benchmark, averaged over three runs. W/T/L counts are summed across runs and shown from the \mqfourb{} perspective.}
\label{tab:pairwise-winrate}
\centering
\footnotesize
\begin{tabular}{lrrrcc}
\toprule
Benchmark & \multicolumn{3}{c}{\mqfourb{} outcomes} & \multicolumn{2}{c}{Pairwise score} \\
\cmidrule(lr){2-4}\cmidrule(lr){5-6}
& Win & Tie & Loss & \mqfourb{} & MedGemma-27B-text-it \\
\midrule
AfriMedQA & 640 & 10{,}258 & 832 & 49.1816\% & \bestcell{50.8184\%} \\
MMLU-Health & 147 & 2{,}944 & 176 & 49.5562\% & \bestcell{50.4438\%} \\
MedQA-USMLE & 325 & 3{,}209 & 285 & \bestcell{50.5237\%} & 49.4763\% \\
PubMedQA & 123 & 1{,}298 & 79 & \bestcell{51.4667\%} & 48.5333\% \\
MedMCQA & 1{,}155 & 10{,}164 & 1{,}230 & 49.7012\% & \bestcell{50.2988\%} \\
MedXpertQA & 1{,}136 & 5{,}476 & 738 & \bestcell{52.7075\%} & 47.2925\% \\
MMLU-Pro Health & 179 & 2{,}032 & 243 & 48.6960\% & \bestcell{51.3040\%} \\
\midrule
Overall (log-weighted) & -- & -- & -- & \bestcell{50.2501\%} & 49.7499\% \\
\bottomrule
\end{tabular}
\end{table}

\begin{table}[!htbp]
\caption{\textbf{Closed-ended decontamination results.} ``Decontam'' evaluates the same checkpoint after excluding benchmark examples matched against the final synthetic training corpus.}
\label{tab:decontam-results}
\centering
\footnotesize
\setlength{\tabcolsep}{3pt}
\adjustbox{max width=\linewidth}{%
\begin{tabular}{llcccccccc}
\toprule
\textbf{Model} & \textbf{Eval set} & \textbf{Avg} & \textbf{AfriMedQA} & \textbf{MMLU-H} & \textbf{MedQA} & \textbf{PubMedQA} & \textbf{MedMCQA} & \textbf{MedXpertQA} & \textbf{MMLU-Pro-H} \\
\midrule
\mqfourb{} & Original & 70.54 & 71.50 & 89.70 & 84.39 & 75.00 & 72.15 & 30.61 & 70.45 \\
\mqfourb{} & Decontam & 70.36 & 71.44 & 88.59 & 84.36 & 75.00 & 72.17 & 30.52 & 70.42 \\
\midrule
MedGemma-27B-text-it & Original & 69.95 & 73.07 & 90.48 & 83.29 & 71.93 & 72.77 & 25.18 & 72.94 \\
MedGemma-27B-text-it & Decontam & 69.80 & 73.01 & 89.70 & 83.25 & 71.93 & 72.79 & 24.98 & 72.95 \\
\midrule
Qwen3-4B-Thinking-2507 & Original & 63.10 & 64.12 & 85.92 & 70.91 & 74.53 & 61.78 & 16.69 & 67.73 \\
Qwen3-4B-Thinking-2507 & Decontam & 62.84 & 64.05 & 84.31 & 70.84 & 74.53 & 61.80 & 16.66 & 67.69 \\
\midrule
\mqonesevenb{} & Original & 62.62 & 64.84 & 82.72 & 75.05 & 69.53 & 63.56 & 21.28 & 61.37 \\
\mqonesevenb{} & Decontam & 62.39 & 64.75 & 81.45 & 74.99 & 69.53 & 63.56 & 21.11 & 61.32 \\
\midrule
MedGemma-1.5-4B-it & Original & 51.20 & 54.38 & 67.69 & 64.39 & 58.73 & 50.08 & 15.80 & 47.31 \\
MedGemma-1.5-4B-it & Decontam & 50.74 & 54.26 & 64.78 & 64.38 & 58.73 & 50.06 & 15.70 & 47.25 \\
\midrule
Qwen3-1.7B (Thinking) & Original & 49.95 & 51.87 & 72.49 & 47.18 & 72.33 & 49.14 & 11.60 & 45.07 \\
Qwen3-1.7B (Thinking) & Decontam & 49.58 & 51.80 & 70.12 & 47.06 & 72.33 & 49.16 & 11.59 & 45.00 \\
\bottomrule
\end{tabular}}
\end{table}

\begin{table}[!htbp]
\caption{\textbf{Decontamination deltas and overlap counts.} Deltas are decontaminated score minus original score.}
\label{tab:decontam-deltas}
\centering
\footnotesize
\setlength{\tabcolsep}{3pt}
\adjustbox{max width=\linewidth}{%
\begin{tabular}{lcccccccc}
\toprule
\textbf{Model / count} & \textbf{Avg} & \textbf{AfriMedQA} & \textbf{MMLU-H} & \textbf{MedQA} & \textbf{PubMedQA} & \textbf{MedMCQA} & \textbf{MedXpertQA} & \textbf{MMLU-Pro-H} \\
\midrule
\mqfourb{} & $-$0.18 & $-$0.06 & $-$1.11 & $-$0.03 & 0.00 & 0.02 & $-$0.09 & $-$0.03 \\
MedGemma-27B-text-it & $-$0.15 & $-$0.06 & $-$0.78 & $-$0.04 & 0.00 & 0.02 & $-$0.20 & 0.01 \\
Qwen3-4B-Thinking-2507 & $-$0.26 & $-$0.07 & $-$1.61 & $-$0.07 & 0.00 & 0.02 & $-$0.03 & $-$0.04 \\
\mqonesevenb{} & $-$0.23 & $-$0.09 & $-$1.27 & $-$0.06 & 0.00 & 0.00 & $-$0.17 & $-$0.05 \\
MedGemma-1.5-4B-it & $-$0.46 & $-$0.12 & $-$2.91 & $-$0.01 & 0.00 & $-$0.02 & $-$0.10 & $-$0.06 \\
Qwen3-1.7B (Thinking) & $-$0.38 & $-$0.07 & $-$2.37 & $-$0.12 & 0.00 & 0.02 & $-$0.01 & $-$0.07 \\
\midrule
Samples removed & 157 & 10 & 125 & 3 & 0 & 1 & 17 & 1 \\
Total count & 14{,}223 & 3{,}910 & 1{,}089 & 1{,}273 & 500 & 4{,}183 & 2{,}450 & 818 \\
Contamination \% & 1.10\% & 0.26\% & 11.48\% & 0.24\% & 0.00\% & 0.02\% & 0.69\% & 0.12\% \\
\bottomrule
\end{tabular}}
\end{table}

\subsection{Quantization: imatrix ablation and per-benchmark tables}
\label{app:quant}

This appendix supplements the six-format quantization study of Sec.~\ref{sec:quant}, whose per-variant overview is main-paper Table~\ref{tab:quant-variants}, with the \texttt{Q4\_K\_M} imatrix ablation, the capacity-vs-precision comparison, and per-benchmark closed-ended scores. The comparison uses \texttt{Q8\_0} without imatrix and all sub-8-bit variants (\texttt{Q4\_K\_M}, \texttt{IQ4\_NL}, \texttt{IQ4\_XS}, \texttt{IQ3\_M}, \texttt{IQ3\_XXS}) with imatrix calibration.

On the 1.7B model, imatrix calibration provides its largest closed-ended benefits on the most reasoning-intensive tasks: 4.69 points on MedQA-USMLE, 3.59 on MMLU-Pro Health, and 2.07 on MMLU (Health), the same benchmarks that benefited most from post-training (Sec.~\ref{subsec:stages}).

\begin{table}[!htbp]
\caption{\textbf{Per-format AVG Score changes by model size} (absolute change vs.\ BF16, on the 0--100 scale).}
\label{tab:quant-cap}
\centering
\footnotesize
\setlength{\tabcolsep}{3pt}
\begin{tabular}{lccc}
\toprule
\textbf{Format} & \textbf{$\Delta$ Score 4B} & \textbf{$\Delta$ Score 1.7B} & \textbf{1.7B / 4B ratio} \\
\midrule
\texttt{Q8\_0}     & $-$0.15 & 0.00 & --- \\
\texttt{Q4\_K\_M}  & $-$0.81 & $-$0.73 & $\approx$1$\times$ \\
\texttt{IQ4\_NL}   & $-$1.02 & $-$1.70 & 1.7$\times$ \\
\texttt{IQ4\_XS}   & $-$1.08 & $-$1.79 & 1.7$\times$ \\
\texttt{IQ3\_M}    & $-$1.50 & $-$3.58 & 2.4$\times$ \\
\texttt{IQ3\_XXS}  & $-$5.56 & $-$12.46 & 2.2$\times$ \\
\bottomrule
\end{tabular}
\end{table}

\begin{table}[!htbp]
\caption{Closed-ended per-benchmark results, quantized \mqfourb{} variants. $\Delta$ vs.\ BF16. ``(+imatrix)'' / ``(no imatrix)'' denote with / without imatrix calibration.}
\label{tab:quant-4b-closed}
\centering
\footnotesize
\setlength{\tabcolsep}{3pt}
\adjustbox{max width=\linewidth}{%
\begin{tabular}{lcccccccccc}
\toprule
Variant & Avg & $\Delta$ & AfriMedQA & MMLU-H & MedQA & PubMedQA & MedMCQA & MedXpertQA & MMLU-Pro-H \\
\midrule
\textbf{4B (BF16)} & \textbf{70.54} & n/a & 71.50 & 89.70 & 84.39 & 75.00 & 72.15 & 30.61 & 70.45 \\
4B-Q8\_0 (+imatrix) & 70.33 & $-$0.21 & 71.15 & 89.19 & 84.57 & 74.73 & 71.82 & 30.05 & 70.82 \\
4B-Q8\_0 (no imatrix) & 70.25 & $-$0.29 & 71.42 & 89.70 & 84.31 & 73.60 & 71.95 & 30.01 & 70.78 \\
4B-Q4\_K\_M (+imatrix) & 69.92 & $-$0.62 & 71.62 & 89.03 & 83.13 & 76.07 & 71.02 & 28.90 & 69.64 \\
4B-Q4\_K\_M (no imatrix) & 69.60 & $-$0.94 & 71.15 & 88.73 & 83.09 & 74.93 & 70.57 & 28.79 & 69.92 \\
\bottomrule
\end{tabular}}
\end{table}

\begin{table}[!htbp]
\caption{HealthBench per-dimension results, quantized \mqfourb{} variants (CompassJudger).}
\label{tab:quant-4b-hb}
\centering
\footnotesize
\setlength{\tabcolsep}{3pt}
\adjustbox{max width=\linewidth}{%
\begin{tabular}{lcccccccc}
\toprule
Variant & Overall & Expert.\ Comm. & Resp.\ Depth & Context Seek. & Emerg.\ Refer. & Global Health & Health Data & Resp.\ Uncert. \\
\midrule
\textbf{4B (BF16)} & \textbf{74.00} & 79.00 & 63.00 & 72.00 & 82.00 & 74.00 & 60.00 & 76.00 \\
4B-Q8\_0 (+imatrix) & 74.00 & 79.00 & 63.00 & 72.00 & 82.00 & 73.00 & 62.00 & 76.00 \\
4B-Q8\_0 (no imatrix) & 74.00 & 79.00 & 62.00 & 72.00 & 82.00 & 74.00 & 60.00 & 76.00 \\
4B-Q4\_K\_M (+imatrix) & 73.00 & 79.00 & 62.00 & 72.00 & 81.00 & 72.00 & 60.00 & 76.00 \\
4B-Q4\_K\_M (no imatrix) & 73.00 & 79.00 & 62.00 & 71.00 & 81.00 & 72.00 & 59.00 & 76.00 \\
\bottomrule
\end{tabular}}
\end{table}

\begin{table}[!htbp]
\caption{Closed-ended per-benchmark results, quantized \mqonesevenb{} variants. $\Delta$ vs.\ BF16.}
\label{tab:quant-1-7b-closed}
\centering
\footnotesize
\setlength{\tabcolsep}{3pt}
\adjustbox{max width=\linewidth}{%
\begin{tabular}{lcccccccccc}
\toprule
Variant & Avg & $\Delta$ & AfriMedQA & MMLU-H & MedQA & PubMedQA & MedMCQA & MedXpertQA & MMLU-Pro-H \\
\midrule
\textbf{1.7B (BF16)} & \textbf{62.62} & n/a & 64.84 & 82.72 & 75.05 & 69.53 & 63.56 & 21.28 & 61.37 \\
1.7B-Q8\_0 (no imatrix) & 62.62 & 0.00 & 64.89 & 82.52 & 74.84 & 70.47 & 63.05 & 21.01 & 61.53 \\
1.7B-Q8\_0 (+imatrix) & 62.44 & $-$0.18 & 64.54 & 82.41 & 74.50 & 71.33 & 62.83 & 21.08 & 60.39 \\
1.7B-Q4\_K\_M (+imatrix) & 62.16 & $-$0.46 & 64.40 & 82.15 & 74.00 & 71.20 & 62.72 & 20.16 & 60.52 \\
1.7B-Q4\_K\_M (no imatrix) & 60.58 & $-$2.04 & 63.24 & 80.65 & 70.36 & 70.67 & 61.22 & 20.12 & 57.78 \\
\bottomrule
\end{tabular}}
\end{table}

\begin{table}[!htbp]
\caption{HealthBench per-dimension results, quantized \mqonesevenb{} variants (CompassJudger).}
\label{tab:quant-1-7b-hb}
\centering
\footnotesize
\setlength{\tabcolsep}{3pt}
\adjustbox{max width=\linewidth}{%
\begin{tabular}{lcccccccc}
\toprule
Variant & Overall & Expert.\ Comm. & Resp.\ Depth & Context Seek. & Emerg.\ Refer. & Global Health & Health Data & Resp.\ Uncert. \\
\midrule
\textbf{1.7B (BF16)} & \textbf{70.00} & 77.00 & 56.00 & 69.00 & 80.00 & 68.00 & 58.00 & 74.00 \\
1.7B-Q8\_0 (no imatrix) & 70.00 & 77.00 & 57.00 & 69.00 & 80.00 & 69.00 & 57.00 & 73.00 \\
1.7B-Q8\_0 (+imatrix) & 70.00 & 77.00 & 57.00 & 70.00 & 79.00 & 69.00 & 56.00 & 74.00 \\
1.7B-Q4\_K\_M (+imatrix) & 69.00 & 75.00 & 56.00 & 69.00 & 79.00 & 67.00 & 55.00 & 73.00 \\
1.7B-Q4\_K\_M (no imatrix) & 68.00 & 74.00 & 54.00 & 67.00 & 78.00 & 66.00 & 55.00 & 72.00 \\
\bottomrule
\end{tabular}}
\end{table}

The per-variant overall scores for the I-quant variants (\texttt{IQ4\_NL}, \texttt{IQ4\_XS}, \texttt{IQ3\_M}, \texttt{IQ3\_XXS}) are reported in main-paper Table~\ref{tab:quant-variants}; Table~\ref{tab:quant-cap} compares score changes by model size.
The imatrix ablation details are split by model size and evaluation type: closed-ended results for 4B and 1.7B are in Tables~\ref{tab:quant-4b-closed} and~\ref{tab:quant-1-7b-closed}, and HealthBench per-dimension results are in Tables~\ref{tab:quant-4b-hb} and~\ref{tab:quant-1-7b-hb}.

\subsection{Real-device measurement protocol}
\label{app:device}
We benchmarked local llama.cpp inference on an Apple M5 Pro using Metal and on a OnePlus 13 (CPH2649IN, Android 16, Adreno 830) using Android build 9226 with Vulkan partial offload (\texttt{-ngl 20}) and six CPU threads. The Apple prompt set contains 20 MedQA and 10 MedXpertQA-Text questions. The phone measurements use three MedQA prompts per model.

\paragraph{Real-device measurements.} We ran the quantized models locally with llama.cpp on an Apple M5 Pro and a OnePlus 13. Table~\ref{tab:device-apple} reports the Apple M5 Pro measurements over the 30-prompt set: matched \texttt{Q4\_K\_M} runs decode at comparable speed (83.9 vs.\ 82.2 tokens/s), but MedPsy-4B emits 738 rather than 2,727 tokens per answer, reducing median end-to-end time from 34.6 to 8.75 seconds (74.7\%). Table~\ref{tab:device-phone} reports the OnePlus 13 measurements, where both model sizes run fully locally within 4 GB of measured peak RAM. Estimated energy per answer shows the same pattern: at matched \texttt{Q4\_K\_M}, \mqfourb{} uses 232 J against the backbone's 917 J (74.7\% lower). The accuracy of these quantized variants on the full benchmark suite is reported in main-paper Table~\ref{tab:quant-variants}.

\begin{table}[!htbp]
\caption{Apple M5 Pro local-inference measurements over the 30-prompt set (20 MedQA and 10 MedXpertQA-Text). Time per answer is the median; energy per answer is estimated as measured sustained SoC power (powermetrics, 26.5 W) times median end-to-end wall time.}
\label{tab:device-apple}
\centering
\footnotesize
\adjustbox{max width=\linewidth}{%
\begin{tabular}{lrrrrr}
\toprule
Model/quant & Peak RAM & tok/s & tokens/answer & time/answer & energy/answer \\
\midrule
MedPsy-1.7B/\texttt{Q4\_K\_M} & 2.60 GB & 171.0 & 994 & 5.9 s & 156 J \\
MedPsy-4B/\texttt{IQ3\_M} & 3.66 GB & 95.0 & 768 & 8.3 s & 220 J \\
MedPsy-4B/\texttt{Q4\_K\_M} & 4.29 GB & 83.9 & 738 & 8.75 s & 232 J \\
Qwen3-4B-Thinking-2507/\texttt{Q4\_K\_M} & 4.28 GB & 82.2 & 2,727 & 34.6 s & 917 J \\
\bottomrule
\end{tabular}}
\end{table}

\begin{table}[!htbp]
\caption{OnePlus 13 local-inference measurements (three MedQA prompts per model). Both models use \texttt{Q4\_K\_M} GGUF builds with Vulkan partial offload (\texttt{-ngl 20}) and six CPU threads. Energy is from the Android battery charge counter (unplugged device at 49\% charge, Wi-Fi debugging).}
\label{tab:device-phone}
\centering
\footnotesize
\begin{tabular}{lrrrrrr}
\toprule
Model & Peak RAM & tok/s & tokens/answer & TTFT & time/answer & energy/answer \\
\midrule
MedPsy-1.7B & 2.22 GB & 15.27 & 789 & 8.90 s & 55.6 s & 420 J \\
MedPsy-4B & 3.98 GB & 6.63 & 422 & 15.04 s & 105.4 s & 508 J \\
\bottomrule
\end{tabular}
\end{table}

\subsection{Accuracy under explicit thinking budgets}
\label{app:thinking-budget}
\begin{figure}[!htbp]
\centering
\includegraphics[width=0.95\linewidth]{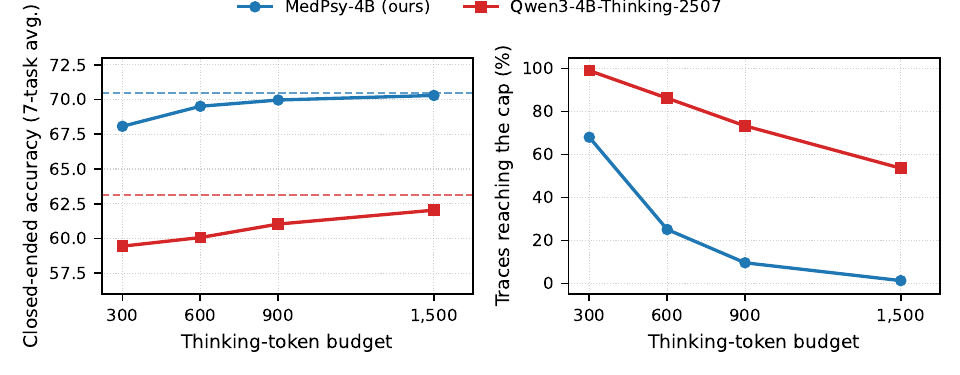}
\caption{\textbf{Accuracy and cap-hit rate under explicit thinking-token caps.} \textbf{Left:} seven-task closed-ended accuracy at thinking budgets of 300--1{,}500 tokens; dashed lines mark unbounded generation. \textbf{Right:} fraction of traces reaching the cap. \mqfourb{} degrades gracefully and rarely reaches the cap beyond a 600-token budget, whereas Qwen3-4B-Thinking-2507 hits the cap on 99\% of traces at 300 tokens and is still cap-bound on over half of traces at 1{,}500.}
\label{fig:thinking-budget}
\end{figure}

We reran the seven closed-ended tasks with external llama.cpp servers while capping tokens in the explicit thinking channel at 300, 600, 900, or 1,500; when the cap was reached, the runtime injected the instruction ``Thinking budget exceeded. Please stop thinking and answer now.'' Each condition contains three sampled generations for all 12,549 examples, evaluated with the QwQ parser. Generation uses thinking mode with temperature 0.6, top-$k$ 20, top-$p$ 0.95, an 8,192-token output limit, and a 16,384-token sequence limit. Table~\ref{tab:thinking-budget} and Figure~\ref{fig:thinking-budget} reports the resulting average accuracy, the fraction of traces that reached the cap, and the mean length of the final answer channel.

\begin{table}[!htbp]
\caption{\textbf{Seven-task closed-ended accuracy under explicit thinking-token caps.} Parentheses show the percentage of traces reaching the cap and mean final-answer-channel tokens.}
\label{tab:thinking-budget}
\centering
\footnotesize
\adjustbox{max width=\linewidth}{%
\begin{tabular}{lccccc}
\toprule
Model / budget & 300 & 600 & 900 & 1,500 & Unbounded \\
\midrule
\mqfourb{} & 68.07 (68.0\%, 152) & 69.51 (25.1\%, 156) & 69.96 (9.6\%, 159) & 70.29 (1.3\%, 163) & 70.48 (0.0\%, 162) \\
Qwen3-4B-Thinking-2507 & 59.45 (99.0\%, 677) & 60.07 (86.2\%, 614) & 61.04 (73.3\%, 578) & 62.04 (53.6\%, 532) & 63.15 (0.0\%, 451) \\
\bottomrule
\end{tabular}}
\end{table}

\mqfourb{} retains 96.6\%, 98.6\%, and 99.3\% of its unbounded score at 300, 600, and 900 thinking tokens; Qwen3-4B-Thinking-2507 reaches 98.2\% at 1,500. At 300 tokens, \mqfourb{} scores 8.62 points above the backbone and 4.92 points above the backbone's unbounded score. These results show stronger accuracy under the tested explicit-reasoning budgets. Qwen3-4B-Thinking-2507 may move some computation into the answer channel under tighter caps, as suggested by its longer final answers.

\subsection{Context-length scaling}
\label{app:context-cost}
On the laptop (Apple M5 Pro, Metal backend), decode throughput at 8K input retains 66.5--72.5\% of its 512-token rate across the three tested variants, while peak RAM at 8K remains between 3.16 and 4.83 GB (Table~\ref{tab:context-laptop}); TTFT at 512 tokens is 0.06--0.12 s. These values come from a dedicated context-scaling probe and differ slightly from the smoke-set figures of Table~\ref{tab:device-apple}; the 4.83 GB peak of \mqfourb{} \texttt{Q4\_K\_M} at 8K context is the largest peak RAM we observed on the laptop.

\begin{table}[!htbp]
\caption{\textbf{Apple M5 Pro context-length sweep} (llama.cpp, Metal): decode throughput at 512-token and 8K-token inputs, peak RAM at 8K, and TTFT at 512 tokens.}
\label{tab:context-laptop}
\centering
\footnotesize
\begin{tabular}{lrrrr}
\toprule
Model/quant & Decode @512 & Decode @8K & Peak RAM @8K & TTFT @512 \\
\midrule
\mqonesevenb{} \texttt{Q4\_K\_M} & 164 tok/s & 109 tok/s & 3.16 GB & 0.06 s \\
\mqfourb{} \texttt{IQ3\_M} & 91 tok/s & 66 tok/s & 4.32 GB & 0.12 s \\
\mqfourb{} \texttt{Q4\_K\_M} & 89 tok/s & 63 tok/s & 4.83 GB & 0.12 s \\
\bottomrule
\end{tabular}
\end{table}

These measurements focus on edge context scaling. Table~\ref{tab:context-phone} reports a OnePlus 13 CPU run with \mqonesevenb{} \texttt{Q4\_K\_M}. At 551 and 4,134 input tokens, TTFT is 12.88 and 93.85 seconds, respectively, while decode throughput is 21.07 and 13.88 tokens/s. These measurements provide practical bounds for selecting context budgets on a phone while process RAM remains stable. This sweep uses the CPU build with six threads; Table~\ref{tab:device-phone} instead reports generation with partial Vulkan offload (\texttt{-ngl 20}).

\begin{table}[!htbp]
\caption{OnePlus 13 CPU context-length scaling with \mqonesevenb{} \texttt{Q4\_K\_M}: timing and throughput across four input lengths.}
\label{tab:context-phone}
\centering
\footnotesize
\adjustbox{max width=\linewidth}{%
\begin{tabular}{lrrrrr}
\toprule
Context target & Input tokens & TTFT (s) & Prefill (tok/s) & Decode (tok/s) & Total time (s) \\
\midrule
512 & 551 & 12.88 & 42.79 & 21.07 & 14.11 \\
1K & 1,062 & 21.23 & 50.04 & 17.82 & 21.95 \\
2K & 2,088 & 51.79 & 40.32 & 14.98 & 53.19 \\
4K & 4,134 & 93.85 & 44.05 & 13.88 & 95.01 \\
\bottomrule
\end{tabular}}
\end{table}

\subsection{Long-context accuracy on LongHealth}
\label{app:longhealth}
LongHealth~\cite{adams2024longhealth} evaluates question answering over long clinical documents (5{,}090--6{,}754 words), probing the input-length regime in which retrieval-augmented deployment operates. Table~\ref{tab:longhealth} reports accuracy on Tasks 1 and 2. Both \medpsy{} sizes outperform the MedGemma model of the next size class up: \mqfourb{} exceeds MedGemma-27B-text-it by 2.7 and 7.3 points on the two tasks, and \mqonesevenb{} exceeds MedGemma-1.5-4B-it by 2.5 and 30.6 points, indicating that the accuracy reported in the main evaluation carries over to long-context inputs.

\begin{table}[!htbp]
\caption{\textbf{LongHealth accuracy} on long clinical documents (5{,}090--6{,}754 words). Best per size class in bold.}
\label{tab:longhealth}
\centering
\footnotesize
\begin{tabular}{lcc}
\toprule
Model & Task 1 & Task 2 \\
\midrule
\mqfourb{} & \bestcell{0.855} & \bestcell{0.875} \\
MedGemma-27B-text-it & 0.828 & 0.802 \\
\midrule
\mqonesevenb{} & \bestcell{0.590} & \bestcell{0.670} \\
MedGemma-1.5-4B-it & 0.565 & 0.364 \\
\bottomrule
\end{tabular}
\end{table}

\subsection{RL ablation studies}
\label{app:ablations}

We ran a controlled set of ablations to isolate which design choices in the \medpsy{} RL recipe drive the gains reported in Tables~\ref{tab:closed}--\ref{tab:hb-judges}. Each ablation modifies a single component of the released \medpsy{} pipeline and trains for the same number of optimisation steps; all other hyperparameters are held constant. The ablations come in two natural groups: (i) \emph{effectiveness} ablations that probe which components contribute to accuracy, and (ii) \emph{token-efficiency} ablations that decompose the source of the response-length compression.

\paragraph{Evaluation suite used in this appendix.}
For every ablation we score checkpoints on the same three closed-ended benchmarks tracked online during training (MedQA-USMLE, MedXpertQA, PubMedQA) using the lightweight flexible scorer in the \texttt{inspect-ai} evaluation framework~\cite{inspectai}. This is a strict subset of the full seven-benchmark suite scored under the LLM-as-Parser methodology of Sec.~\ref{subsec:eval-method} and Table~\ref{tab:tokens-detail}; the smoke suite was used to keep ablation compute and time within budget. Per-checkpoint absolute scores in this appendix therefore differ slightly from the LLM-as-Parser numbers in the main paper. For example, on MedQA-USMLE the released \mqfourb{} S2 checkpoint scores $84.84\%$ under \texttt{inspect-ai} vs.\ $84.39\%$ under LLM-as-Parser, well within scorer noise. Output-token counts are also broadly consistent across the two scorers but can differ in absolute value due to different prompting conventions and post-generation truncation in the two frameworks; the per-ablation \emph{deltas} reproduce on both. We use the smoke suite throughout this appendix and direct readers wanting released-checkpoint absolute numbers to the main-paper tables.

\paragraph{Ablation set.}
All ablations train from the same SFT-Stage-2 checkpoints used by the released \medpsy{} models, except \emph{RL-from-base}, which uses no SFT at all. We use short, descriptive names throughout this appendix:
\begin{itemize}[leftmargin=*, itemsep=2pt, topsep=2pt]
\item \emph{Random S2.} Same Stage-2 setup as \medpsy{} but the training data is a uniformly-random sample from AlphaMed19K~\cite{alphamedqa}, sized to match the mined 1:2 right:wrong subset (2{,}439 items at 4B, 4{,}317 at 1.7B). Tests whether the mining procedure itself contributes vs.\ training longer on more data.
\item \emph{Binary reward.} Replace the four-tier reasoning-aware reward (Eq.~\ref{eq:reward-task}) with the binary $\{0,1\}$ correctness reward used by AlphaMed~\cite{alphamedqa}. Isolates the contribution of the gated reward.
\item \emph{No group filter.} Disable DAPO's dynamic filtering of zero-variance rollout groups (\texttt{filter\_groups.\allowbreak enable=false}). Isolates the contribution of group filtering.
\item \emph{No overlong.} Remove the soft penalty applied to responses exceeding $L_{\text{soft}}{=}2{,}048$ tokens within the $L_{\text{max}}{=}4{,}096$-token budget; all other DAPO settings unchanged. Isolates the contribution of the soft length penalty.
\item \emph{Vanilla GRPO.} Replace DAPO with vanilla GRPO~\cite{grpo}: re-introduce the KL anchor (\texttt{kl\_coef}$=0.001$), use symmetric PPO clipping ($\epsilon_{\text{low}}{=}\epsilon_{\text{high}}{=}0.2$), and disable the overlong penalty. Isolates the contribution of DAPO's full set of drift-enabling modifications.
\item \emph{RL-from-base.} RL Stage 1 initialised from the un-fine-tuned Qwen3 backbones (Qwen3-4B-Thinking-2507 for the 4B class, Qwen3-1.7B for the 1.7B class), with no SFT at all. Tests whether SFT is necessary before RL.
\item \emph{Difficult-only S1.} RL Stage 1 trained only on items the size-matched SFT-2 checkpoint fails on every attempt during pass-at-$N$ annotation ($n_c{=}0$); $1{,}276$ items at 4B and $1{,}888$ items at 1.7B. Tests whether training on easy/moderate items is necessary to install short reasoning.
\end{itemize}

\subsubsection{Effectiveness ablations}
\label{subsec:abl-effectiveness}

\paragraph{Mining drives hard-case generalisation.}
The clearest effectiveness finding is that the deterministic 1:2 right:wrong mining of Stage 2 (\S\ref{para:mining}) accounts for the majority of the Stage-2 lift, and it does so almost entirely through gains on the hardest of our benchmarks (MedXpertQA). Holding the Stage-1 checkpoint and all hyperparameters fixed (Table~\ref{tab:abl-mining}), the mined Stage-2 set lifts MedXpertQA by $+6.1$~pp on the 4B model and $+4.3$~pp on the 1.7B model, whereas the \emph{Random S2} ablation lifts it by only $+1.1$~pp and $+2.3$~pp respectively. Easy and medium benchmarks (MedQA, PubMedQA) are essentially indistinguishable between mined and random Stage-2 (within the per-benchmark stderr of $\sim 1.0$~pp). The mined Stage-2 also reaches its peak in fewer steps: for the 4B model the mined run peaks at step $190$, while the random run requires the full $500$ steps to reach a lower plateau.

\begin{table}[!htbp]
\caption{\textbf{Mined vs.\ random Stage-2.} Both runs initialise from the same Stage-1 best checkpoint and train for the same number of steps. Mining's benefit is concentrated on MedXpertQA at both scales.}
\label{tab:abl-mining}
\centering
\footnotesize
\setlength{\tabcolsep}{3pt}
\adjustbox{max width=\linewidth}{%
\begin{tabular}{l c c c c c c}
\toprule
& \textbf{Best step} & \textbf{MedQA} & \textbf{MedXpertQA} & \textbf{PubMedQA} & \textbf{Avg} & \textbf{$\Delta$ vs.\ S1} \\
\midrule
\multicolumn{7}{l}{\emph{4B class (S1 best avg = 62.34)}}\\
S2 mined (\medpsy{})        & 190 & 84.76 & \bestcell{32.37} & 76.40 & \bestcell{64.51} & +2.17 \\
S2 random (\emph{Random S2})  & 500 & 85.31 & 27.39 & 77.00 & 63.23 & +0.89 \\
\midrule
\multicolumn{7}{l}{\emph{1.7B class (S1 best avg = 52.35)}}\\
S2 mined (\medpsy{})        & 400 & 72.82 & \bestcell{20.94} & 76.60 & \bestcell{56.79} & +4.44 \\
S2 random (\emph{Random S2})  & 230 & 74.71 & 18.90 & 73.40 & 55.67 & +3.32 \\
\bottomrule
\end{tabular}
}
\end{table}

This pattern matches the design intent: the 1:2 right:wrong subset over-samples failure cases that, by construction, require capabilities beyond surface pattern matching. MedXpertQA requires advanced reasoning patterns to pass; MedQA and PubMedQA are dominated by simpler factual recall and structured-text comprehension, which both Stage-1 RL and even random Stage-2 already saturate.

\paragraph{Reward shape and DAPO group filter are within smoke-eval noise.}
The \emph{Binary reward} ablation, which replaces the four-tier reasoning-aware reward with AlphaMed's binary $\{0,1\}$ correctness reward, leaves average accuracy within $\pm 1.0$~pp of the \medpsy{} Stage-1 best at both model sizes. The \emph{No group filter} ablation, which disables DAPO's dynamic group filtering, is similarly within $\pm 1.0$~pp (Table~\ref{tab:abl-other}). We do not claim an accuracy contribution from either knob based on these results. The four-tier reward does, however, produce a small but measurable token-efficiency benefit relative to binary, which we attribute to the explicit penalisation of the ``correct without thinking'' shortcut.\footnote{For the 4B model, the average MedQA output tokens at the best checkpoint are $759$ for \medpsy{} Stage 1 (4-tier reward) vs.\ $836$ for the \emph{Binary reward} ablation, about $9\%$ lower for the four-tier reward at matched accuracy. For the 1.7B model the gap is within noise ($943$ vs.\ $944$ tokens).}

\paragraph{RL needs SFT.}
The \emph{RL-from-base} ablation initialises RL Stage 1 from the un-fine-tuned Qwen3 backbones (Qwen3-4B-Thinking-2507 and Qwen3-1.7B), bypassing both SFT stages. Performance plateaus far below the \medpsy{} Stage 1 at both model sizes (Table~\ref{tab:abl-k}): the 4B \emph{RL-from-base} run reaches a smoke-suite average of $55.89\%$ vs.\ $62.34\%$ for the \medpsy{} Stage 1 ($-6.45$~pp), and the 1.7B \emph{RL-from-base} run reaches $45.07\%$ vs.\ $52.35\%$ ($-7.28$~pp). Two observations clarify the gap. First, the absolute RL lift over the run's starting point is comparable to the \medpsy{} recipe: the 4B \emph{RL-from-base} run improves from $54.20\%$ at step $0$ to $55.89\%$ at its best step ($+1.69$~pp), while \medpsy{} Stage 1 improves from $60.69\%$ to $62.34\%$ ($+1.65$~pp); both gain $\sim 1.7$~pp from RL. What differs is the floor, which sits $6$--$7$~pp lower without SFT. Second, the training reward signal in the \emph{RL-from-base} runs is dominated by missing format compliance: at the 4B size the mean reward converges to $\sim 0.43$ vs.\ $\sim 0.96$ for \medpsy{}, with the gap explained by the model failing to reliably emit the structured \texttt{<think>...</think>}/\texttt{<answer>...</answer>} format that the SFT pipeline installs (Eq.~\ref{eq:reward-task} returns $0.5$ instead of $1.0$ even when the answer is correct). We therefore conclude that SFT establishes the operating floor -- broad medical coverage and the reasoning-trace format -- on top of which RL adds its $\sim 1.7$~pp of consolidation; RL on the un-fine-tuned backbone is not unable to learn, but cannot recover the SFT-installed capabilities in the same number of optimisation steps. Per-benchmark numbers in Table~\ref{tab:abl-k} are consistent with the corresponding backbone numbers reported under LLM-as-Parser in Sec.~\ref{subsec:stages} (e.g.\ MedQA-USMLE on the 4B Qwen3 backbone is $71.56\%$ under LLM-as-Parser vs.\ $72.19\%$ for \emph{RL-from-base} baseline under \texttt{inspect-ai}; MedQA-USMLE on the 1.7B Qwen3 backbone is $47.18\%$ under LLM-as-Parser vs.\ $47.13\%$ under \texttt{inspect-ai}).

\begin{table}[!htbp]
\caption{\textbf{\emph{RL-from-base} ablation: RL Stage 1 from the un-fine-tuned Qwen3 backbones.} Best-step on the smoke-eval suite. \emph{RL-from-base} converges $6$--$7$~pp below the \medpsy{} Stage 1 at both model sizes because it lacks the SFT-installed medical coverage and reasoning-trace format, even though the absolute RL lift over its starting point is comparable to \medpsy{}.}
\label{tab:abl-k}
\centering
\footnotesize
\setlength{\tabcolsep}{3pt}
\adjustbox{max width=\linewidth}{%
\begin{tabular}{l c c c c c c}
\toprule
& \textbf{Best step} & \textbf{MedQA} & \textbf{MedXpertQA} & \textbf{PubMedQA} & \textbf{Avg} & \textbf{$\Delta$ vs.\ \medpsy{} S1} \\
\midrule
\multicolumn{7}{l}{\emph{4B class}}\\
\medpsy{} S1 (from SFT-2)               & 112 & 85.70 & 26.33 & 75.00 & \bestcell{62.34} & --- \\
\emph{RL-from-base} (no SFT, baseline)    &   0 & 72.19 & 18.20 & 72.20 & 54.20 & $-$8.14 \\
\emph{RL-from-base} (no SFT, best)        & 120 & 74.47 & 18.41 & 74.80 & 55.89 & $-$6.45 \\
\midrule
\multicolumn{7}{l}{\emph{1.7B class}}\\
\medpsy{} S1 (from SFT-2)               &  96 & 69.29 & 16.57 & 71.20 & \bestcell{52.35} & --- \\
\emph{RL-from-base} (no SFT, baseline)    &   0 & 47.13 & 11.76 & 73.00 & 43.96 & $-$8.39 \\
\emph{RL-from-base} (no SFT, best)        &  68 & 49.18 & 12.24 & 73.80 & 45.07 & $-$7.28 \\
\bottomrule
\end{tabular}
}
\end{table}

\begin{table}[!htbp]
\caption{\textbf{\emph{Binary reward} and \emph{No group filter} ablations on RL Stage 1.} All numbers are best-step on the smoke-eval suite. Both ablations are within smoke-eval noise of the \medpsy{} Stage-1 best at both model sizes; we report them as \emph{neutral} accuracy findings.}
\label{tab:abl-other}
\centering
\footnotesize
\setlength{\tabcolsep}{3pt}
\adjustbox{max width=\linewidth}{%
\begin{tabular}{l c c c c c c}
\toprule
& \textbf{Best step} & \textbf{MedQA} & \textbf{MedXpertQA} & \textbf{PubMedQA} & \textbf{Avg} & \textbf{$\Delta$ vs.\ \medpsy{} S1} \\
\midrule
\multicolumn{7}{l}{\emph{4B class (\medpsy{} S1 best = 62.34, MedXpertQA = 26.33)}}\\
\medpsy{} S1 (4-tier $+$ filter $+$ overlong) & 112 & 85.70 & 26.33 & 75.00 & 62.34 & --- \\
\emph{Binary reward}                            & 84  & 84.92 & 25.55 & 74.60 & 61.69 & $-$0.65 \\
\emph{No group filter}                          & 124 & 85.07 & 26.33 & 75.60 & 62.33 & $\pm$0.00 \\
\midrule
\multicolumn{7}{l}{\emph{1.7B class (\medpsy{} S1 best = 52.35, MedXpertQA = 16.57)}}\\
\medpsy{} S1 (4-tier $+$ filter $+$ overlong) & 96  & 69.29 & 16.57 & 71.20 & 52.35 & --- \\
\emph{Binary reward}                            & 100 & 69.13 & 17.18 & 71.80 & 52.70 & $+$0.35 \\
\emph{No group filter}                          & 40  & 66.38 & 16.65 & 74.60 & 52.54 & $+$0.19 \\
\bottomrule
\end{tabular}
}
\end{table}

\subsubsection{Token-efficiency ablations: a clean three-way decomposition}
\label{subsec:abl-tokenefficiency}

The $1.7$--$3.2\times$ token-efficiency improvements of the \medpsy{} models reported in Sec.~\ref{sec:tokeneff} are the most distinctive empirical property of our pipeline. The \emph{Vanilla GRPO} and \emph{No overlong} ablations, in combination with the \medpsy{} runs, decompose this compression into three additive sources, each separately controllable from the others. Table~\ref{tab:abl-tokens} reports per-benchmark average output tokens at the best checkpoint of each variant; Table~\ref{tab:abl-compress} reports the corresponding training-time response-length trajectory.

\paragraph{Three additive sources.}
Reading Table~\ref{tab:abl-tokens} top-to-bottom within each model class:
\begin{itemize}[leftmargin=*, itemsep=2pt, topsep=2pt]
\item \textbf{(1) RL with a KL anchor (\emph{Vanilla GRPO}).} Vanilla GRPO compresses MedQA tokens by only $6\%$ on the 4B model and $20\%$ on the 1.7B model relative to the SFT-Stage-2 baseline. The KL anchor pulls the policy back toward the verbose SFT prior, and symmetric clipping limits how far the upper PPO bound can boost short-but-correct trajectories. The accuracy delta of \emph{Vanilla GRPO} is also small ($-0.5$~pp on the 4B model, $-0.6$~pp on the 1.7B model vs.\ the \medpsy{} Stage 1).
\item \textbf{(2) DAPO drift modifications (\emph{No overlong}).} Switching to DAPO, removing the KL term and using clip-higher$=0.28$, without the overlong penalty pushes compression to $44\%$ (4B) and $45\%$ (1.7B). Allowing the policy to drift further from the SFT prior and biasing the upper PPO bound is responsible for the bulk of the compression, even though neither change references response length explicitly. This ablation also delivers the highest accuracy of any 4B Stage-1 variant ($+1.0$~pp vs.\ \medpsy{}).
\item \textbf{(3) The soft overlong-response penalty (\medpsy{}).} Adding the linear penalty over $[L_{\text{soft}}, L_{\text{max}}]$ on top of DAPO drives compression the rest of the way to $66\%$ (4B) and $65\%$ (1.7B). The overlong penalty contributes its $\sim 22$~pp of additional compression with a measured accuracy cost of $1.0$~pp on the 4B model and $0.8$~pp on the 1.7B model relative to \emph{No overlong}; it is also the only component that prevents the right tail of the response-length distribution from re-expanding (Table~\ref{tab:abl-compress}, ``Tail (end)'' reports the fraction of rollouts at the $4{,}096$-token cap). Removing the penalty leaves the mean below $L_{\text{soft}}$ but allows individual rollouts to grow, increasing average output length at evaluation by approximately $1.6\times$.
\end{itemize}

\begin{table}[!htbp]
\caption{\textbf{Token-efficiency decomposition (output tokens at evaluation).} Per-benchmark average output tokens at the best Stage-1 checkpoint of each variant, plus the final \medpsy{} Stage 2 row for reference. The three drift-side mechanisms (Vanilla GRPO $\to$ DAPO drift mods $\to$ overlong penalty) compose additively to the released-model token budget.}
\label{tab:abl-tokens}
\centering
\footnotesize
\setlength{\tabcolsep}{3pt}
\adjustbox{max width=\linewidth}{%
\begin{tabular}{l c c c c}
\toprule
& \textbf{MedQA} & \textbf{MedXpertQA} & \textbf{PubMedQA} & \textbf{vs.\ SFT-2} \\
\midrule
\multicolumn{5}{l}{\emph{4B class}}\\
SFT-2 baseline (no RL)                          & 2{,}232 & 3{,}510 & 728 & --- \\
{}+ \emph{Vanilla GRPO}                         & 2{,}096 & 3{,}199 & 695 & $-$6\% / $-$9\% / $-$5\% \\
{}+ DAPO drift mods (\emph{No overlong})        & 1{,}259 & 1{,}986 & 555 & $-$44\% / $-$43\% / $-$24\% \\
{}+ overlong penalty (\medpsy{} Stage 1)      & \bestcell{759} & \bestcell{1{,}183} & \bestcell{374} & $-$66\% / $-$66\% / $-$49\% \\
{}+ Stage-2 mining (\mqfourb{} S2)              & 653 & 1{,}066 & 356 & $-$71\% / $-$70\% / $-$51\% \\
\midrule
\multicolumn{5}{l}{\emph{1.7B class}}\\
SFT-2 baseline (no RL)                          & 2{,}674 & 3{,}642 & 938 & --- \\
{}+ \emph{Vanilla GRPO}                         & 2{,}142 & 2{,}792 & 805 & $-$20\% / $-$23\% / $-$14\% \\
{}+ DAPO drift mods (\emph{No overlong})        & 1{,}475 & 2{,}117 & 696 & $-$45\% / $-$42\% / $-$26\% \\
{}+ overlong penalty (\medpsy{} Stage 1)      & \bestcell{943} & \bestcell{1{,}204} & \bestcell{464} & $-$65\% / $-$67\% / $-$51\% \\
{}+ Stage-2 mining (\mqonesevenb{} S2)          & 1{,}024 & 1{,}244 & 539 & $-$62\% / $-$66\% / $-$43\% \\
\bottomrule
\end{tabular}
}
\end{table}

\begin{table}[!htbp]
\caption{\textbf{Training-time response length and tail behaviour by Stage-1 variant.} ``Mean (start)'' and ``Mean (end)'' are per-rollout response lengths averaged over the first/last five logged steps; ``Tail (end)'' is the fraction of rollouts that hit the $4{,}096$-token cap in the final five steps. Removing the overlong penalty keeps the tail wide and prevents the mean from compressing further.}
\label{tab:abl-compress}
\centering
\footnotesize
\setlength{\tabcolsep}{3pt}
\adjustbox{max width=\linewidth}{%
\begin{tabular}{l c c c c c c}
\toprule
& \multicolumn{3}{c}{\textbf{4B class}} & \multicolumn{3}{c}{\textbf{1.7B class}}\\
\cmidrule(lr){2-4} \cmidrule(lr){5-7}
\textbf{Stage-1 variant}
& \textbf{Mean (start)} & \textbf{Mean (end)} & \textbf{Tail (end)}
& \textbf{Mean (start)} & \textbf{Mean (end)} & \textbf{Tail (end)} \\
\midrule
\medpsy{} (DAPO + overlong)             & 1{,}128 & 633   & 0.00\% & 1{,}270 & 787   & 0.06\% \\
DAPO, \emph{No overlong}                  & 1{,}482 & 1{,}154 & 0.16\% & 1{,}556 & 1{,}220 & 0.74\% \\
\emph{Vanilla GRPO} (KL + symmetric clip) & 1{,}762 & 1{,}699 & 3.60\% & 1{,}901 & 1{,}760 & 3.99\% \\
\bottomrule
\end{tabular}
}
\end{table}

\paragraph{The accuracy/efficiency trade-off is modest.}
A common concern with length-controlled RL is that compression comes at the cost of accuracy on hard items. In our setting, the measured cost is at most approximately one point on the smoke suite. The \emph{No overlong} ablation, which keeps DAPO but disables the soft length penalty, scores $63.34\%$ avg on the 4B model (vs.\ $62.34\%$ for the \medpsy{} Stage 1, $+1.00$~pp) at the cost of $\sim 1.6\times$ longer outputs, and $53.15\%$ on the 1.7B model (vs.\ $52.35\%$, $+0.80$~pp), again at $\sim 1.6\times$ length. The overlong penalty therefore buys the headline $1.7$--$3.2\times$ token reduction reported in Sec.~\ref{sec:tokeneff} for a worst-case accuracy cost of $\sim 1$~pp on the smoke suite. For an on-device deployment this trade-off is decisively in favour of the penalty; for a deployment that prioritises accuracy over latency, dropping the penalty is a defensible alternative.

\paragraph{Why mean tokens shrink even when the penalty is rarely triggered.}
A natural objection to the overlong-penalty story is that the policy's mean response length is well below the $L_{\text{soft}}{=}2{,}048$-token threshold for almost the entire run, so the penalty should ``not matter''. Table~\ref{tab:abl-compress} resolves this: the penalty does not act on the mean, but on the per-rollout right tail. In the \medpsy{} runs, the fraction of rollouts that hit the $L_{\text{max}}{=}4{,}096$-token cap drops from $2$--$3\%$ in the first decile of training to $\sim 0$--$0.1\%$ at the end. Removing the penalty (the \emph{No overlong} ablation) keeps that tail wide ($0.16$--$0.74\%$ of rollouts at the cap), and the wider tail in turn pulls the mean upward via DAPO's drift dynamics. The penalty is best understood as an \emph{anchor on the right tail of the response-length distribution}, not as a per-step length subtraction.

\begin{figure}[!htbp]
\centering
\includegraphics[width=0.95\linewidth]{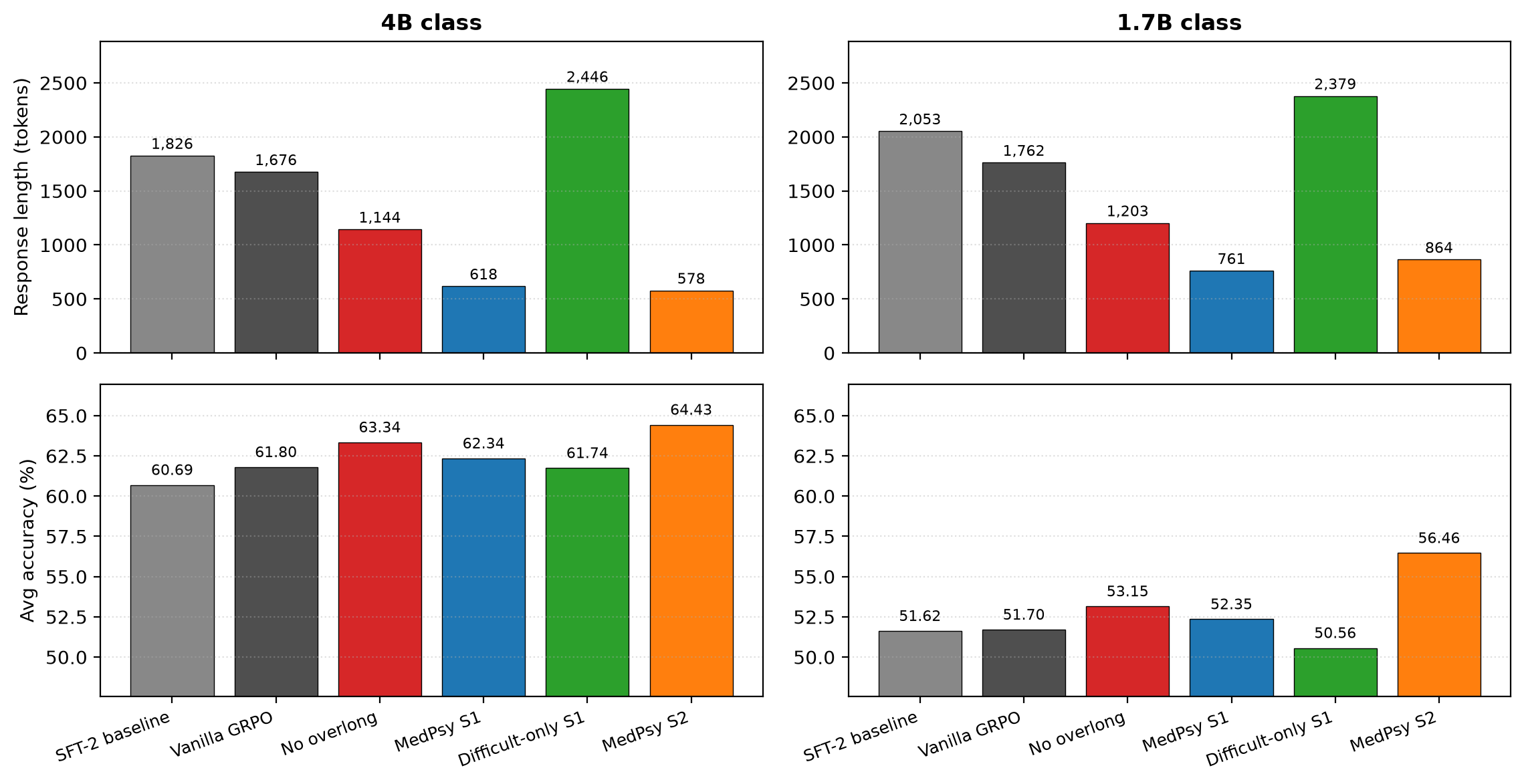}
\caption{\textbf{Training-time mean response length (top) and avg accuracy (bottom) at the best-accuracy checkpoint of each RL variant.} \medpsy{} S2 (orange) wins on both axes at both model sizes.}
\label{fig:rl-summary-bars}
\end{figure}

\begin{figure}[!htbp]
\centering
\includegraphics[width=0.95\linewidth]{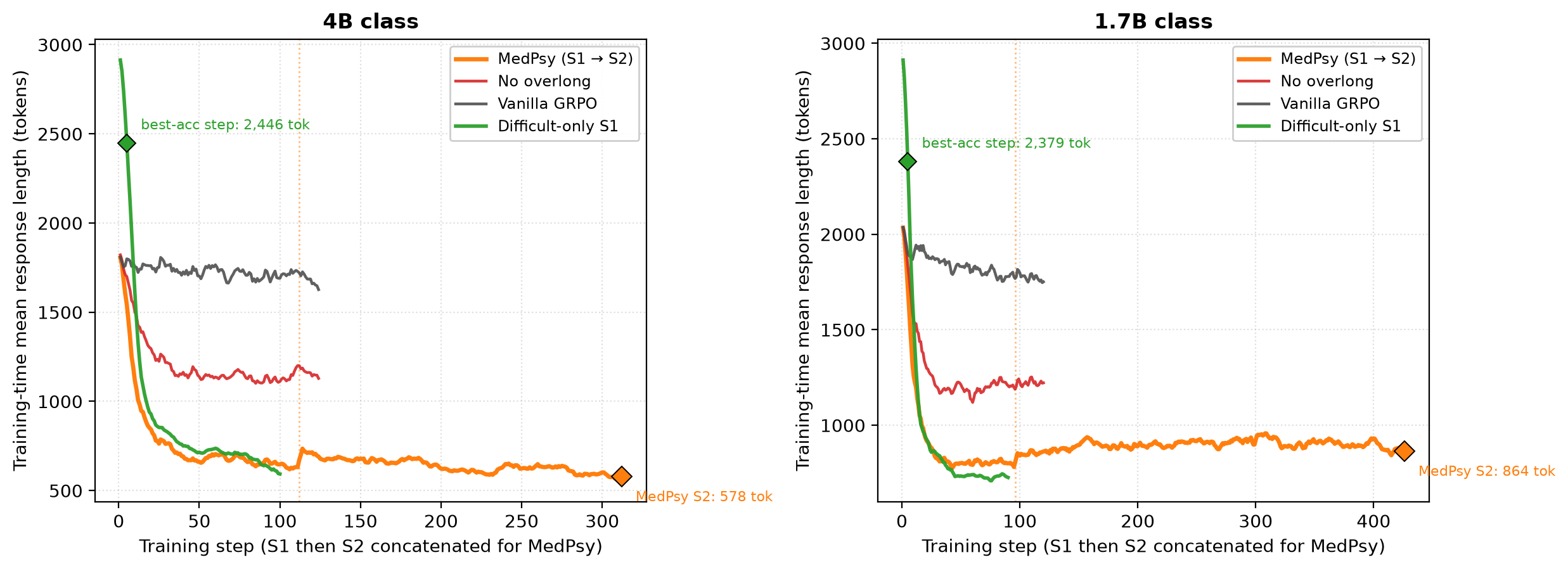}
\caption{\textbf{Training-time mean response length over training, traces truncated at the best-accuracy checkpoint.} \medpsy{} S2 (orange) traces the full pipeline: Easy/Moderate Stage~1 compresses the policy from $\sim 1.8$--$2.0$K tokens down to $\sim 620$--$760$ tokens, then mined-hard Stage~2 (after the dotted vertical line) converges to $\sim 580$--$870$ tokens. The \emph{Difficult-only S1} ablation (green) reaches its best-accuracy checkpoint very early at $\sim 2.4$K tokens (diamond marker), and only matches \medpsy{} S2's response length late in training, after losing $6$--$9$~pp of accuracy in the process. \emph{No overlong} (red) plateaus much higher; \emph{Vanilla GRPO} (gray) barely moves. Diamond markers show the best-accuracy checkpoint values for \medpsy{} S2 and \emph{Difficult-only S1}. }
\label{fig:rl-training-curriculum}
\end{figure}

\begin{figure}[!htbp]
\centering
\includegraphics[width=0.85\linewidth]{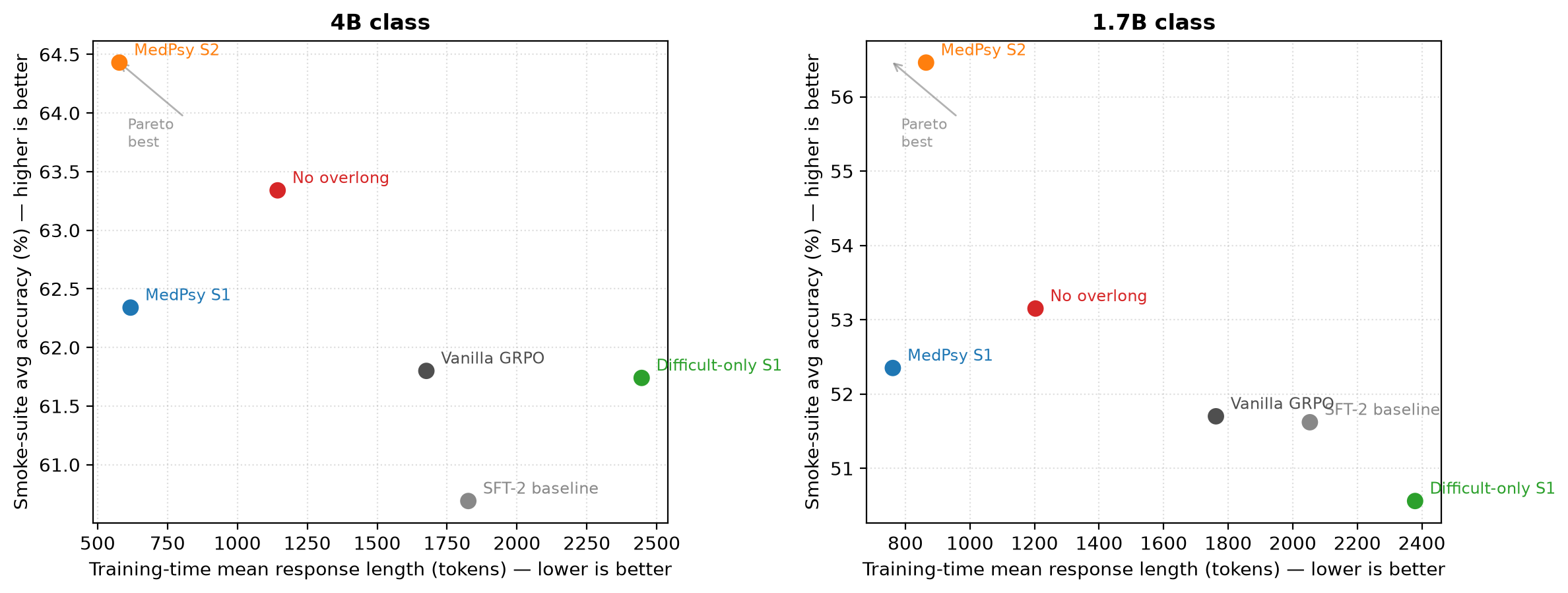}
\caption{\textbf{Token--accuracy Pareto across all RL variants at their best-accuracy checkpoints} (upper-left is better; \medpsy{} S2 is the orange diamond). Response length on the x-axis is the training-time mean at the same checkpoint used for the accuracy on the y-axis. \medpsy{} S2 is the unique Pareto-dominant point at both model sizes: shortest responses and highest accuracy. }
\label{fig:rl-pareto}
\end{figure}

\paragraph{The full curriculum Pareto-strictly-dominates the single-stage Difficult-only shortcut.}
The cleanest counterfactual to the two-stage curriculum is to skip the easy/moderate Stage 1 entirely and run a single RL stage on hard data. The \emph{Difficult-only S1} ablation does exactly this: it initializes from the size-matched SFT-2 checkpoint and trains DAPO on only the items the SFT-2 policy fails on every attempt during pass-at-$N$ annotation ($n_c{=}0$); the same per-policy mining procedure used to construct the released Stage~2 set. Table~\ref{tab:abl-j-detail}, Table~\ref{tab:abl-j-summary}, and Figs.~\ref{fig:rl-summary-bars}--\ref{fig:rl-pareto} compare it to \medpsy{} S2 (\medpsy{} = SFT-2 $\to$ RL Stage 1 on Easy/Moderate $\to$ RL Stage 2 on mined hard). The result is that \emph{every} checkpoint along the \emph{Difficult-only} trajectory is Pareto-dominated by \medpsy{} S2 on accuracy and training-time response length, simultaneously. At the best-by-accuracy checkpoint \medpsy{} S2 is better on \emph{both} axes by a wide margin: $+2.69$~pp average accuracy and $-1{,}868$ training-time tokens on the 4B model ($64.43\%$ / $578$ tokens vs.\ $61.74\%$ / $2{,}446$ tokens), and $+5.90$~pp / $-1{,}515$ tokens on the 1.7B model ($56.46\%$ / $864$ tokens vs.\ $50.56\%$ / $2{,}379$ tokens). On the hardest benchmark MedXpertQA the curriculum's lead widens to $+4.33$~pp on the 4B model ($31.27\%$ vs.\ $26.94\%$) and $+4.82$~pp on the 1.7B model ($21.31\%$ vs.\ $16.49\%$).

\paragraph{Why the curriculum wins, and why \emph{Difficult-only} validates the design.}
Fig.~\ref{fig:rl-training-curriculum} traces the training-time mean response length over the full curriculum (the \medpsy{} trace truncated at the released S2 checkpoint, S2 step $200$ for the 4B model and S2 step $330$ for the 1.7B model). The Easy/Moderate Stage~1 brings the policy from the verbose SFT-2 prior ($\sim 1.8$--$2.0$K tokens) down to $\sim 620$--$760$ tokens; Stage~2 then trains on hard items \emph{starting from} that compressed checkpoint and converges to $\sim 580$--$870$ tokens, gaining accuracy throughout. \emph{Difficult-only S1}, by contrast, sees no items it can solve at the start of training (every prompt has $n_c{=}0$ under the SFT-2 policy by construction), so the available reward gradient is dominated by overlong-length and shortcut penalties rather than correctness signal. As a result the policy compresses tokens at essentially the same per-step rate as all DAPO+overlong runs ($-65$\% to $-77$\% across the run on training-time response length) but \emph{loses} accuracy as it does so: on the 4B model the run peaks at step $4$ ($61.74\%$, still close to the SFT-2 starting accuracy of $60.69\%$) and ends at step $100$ at $58.08\%$; on the 1.7B model it peaks at step $4$ ($50.56\%$) and ends at step $90$ at $47.73\%$. By the end of training, training-time tokens are essentially matched between \emph{Difficult-only S1} ($\sim 590$ on the 4B model, $\sim 730$ on the 1.7B model) and \medpsy{} S2 ($578$ / $864$), but accuracy is now $6$--$9$~pp lower. The trajectory therefore validates the design in a stronger form than we initially anticipated: easy/moderate items are necessary not merely to install short reasoning, but to provide a learnable correctness signal at all -- a single RL stage on the items the same SFT-2 policy never solves cannot make accuracy progress, and the only thing that does happen with more steps is that the soft-overlong penalty compresses tokens at the cost of accuracy. Easy-first followed by mined hard is what simultaneously installs the short-reasoning habit and gains the hard-item accuracy that defines \medpsy{} S2.

\paragraph{A note on units.}
``Response length'' refers to the mean number of tokens the policy emits when it generates a single answer. We report it in two forms that are conceptually the same and differ only in implementation: the \emph{training-time} value is averaged over the $n{=}16$ rollouts per training prompt (sampled at temperature $0.7$ during DAPO group sampling) and is logged every step in W\&B; the \emph{evaluation} value (used in the per-benchmark token tables of Sec.~\ref{sec:tokeneff} and Table~\ref{tab:abl-j-detail}) is computed over the standard MedQA / MedXpertQA / PubMedQA prompts under deterministic generation. Both views agree on the headline finding (\medpsy{} S2 Pareto-dominates the \emph{Difficult-only} trajectory); we use training-time in Figs.~\ref{fig:rl-training-curriculum}--\ref{fig:rl-summary-bars} because it is per-step granular and free of evaluation-prompt confounds.

\paragraph{Would a longer \emph{Difficult-only} run catch up?}
\emph{Difficult-only S1} is trained for $50$ epochs ($\sim 100$ optimisation steps) on the 4B model and $30$ epochs ($\sim 90$ steps) on the 1.7B model, matched to the \medpsy{} Stage 1 step count, with each model trained on the items its \emph{own} size-matched SFT-2 policy fails on every $n{=}5$ rollout ($1{,}276$ items at 4B, $1{,}888$ at 1.7B). Within the tested budget, two signals argue against near-term catch-up, while the smaller and less diverse training set remains a confound. First, the accuracy trajectory degrades after the first few steps (from $61.74\%$ at step $4$ down to $\sim 58$\% on the 4B model; from $50.56\%$ at step $4$ down to $\sim 48$\% on the 1.7B model). Second, the training-time response length is approximately saturated by step $\sim 80$ at both sizes ($\sim 590$--$650$ tokens on the 4B model, $\sim 700$--$750$ tokens on the 1.7B model). The observed \emph{Difficult-only} trajectory is therefore dominated by \medpsy{} S2 within this training budget; this does not establish a universal result for data-volume-matched or substantially longer runs.

\begin{table}[!htbp]
\caption{\textbf{Headline ablation summary.} Training-time mean response length at the best-accuracy checkpoint and best-checkpoint avg accuracy on the 3-benchmark smoke suite. For the SFT-2 baseline the response length is the training-rollout length at the first RL step (since SFT-2 has no RL trace itself). \medpsy{} S2 has the shortest response length \emph{and} the highest accuracy at both model sizes. Token reductions are relative to the SFT-2 baseline.}
\label{tab:abl-j-summary}
\centering
\footnotesize
\setlength{\tabcolsep}{3pt}
\adjustbox{max width=\linewidth}{%
\begin{tabular}{l rr rr rr rr}
\toprule
& \multicolumn{4}{c}{\textbf{4B class}} & \multicolumn{4}{c}{\textbf{1.7B class}} \\
\cmidrule(lr){2-5}\cmidrule(lr){6-9}
& \textbf{Tokens} & \textbf{$\downarrow$ vs SFT-2} & \textbf{Acc} & \textbf{$\Delta$ vs SFT-2}
& \textbf{Tokens} & \textbf{$\downarrow$ vs SFT-2} & \textbf{Acc} & \textbf{$\Delta$ vs SFT-2} \\
\midrule
SFT-2 baseline                  & 1{,}826 & ---       & 60.69 & ---            & 2{,}053 & ---       & 51.62 & ---           \\
\emph{Vanilla GRPO}             & 1{,}676 & $-8$\%    & 61.80 & $+1.11$        & 1{,}762 & $-14$\%   & 51.70 & $+0.08$       \\
\emph{No overlong}              & 1{,}144 & $-37$\%   & 63.34 & $+2.65$        & 1{,}203 & $-41$\%   & 53.15 & $+1.53$       \\
\medpsy{} S1                  & 618     & $-66$\%   & 62.34 & $+1.65$        & 761     & $-63$\%   & 52.35 & $+0.73$       \\
\emph{Difficult-only S1}        & 2{,}446 & $+34$\%   & 61.74 & $+1.05$        & 2{,}379 & $+16$\%   & 50.56 & $-1.06$       \\
\textbf{\medpsy{} S2}         & \bestcell{\textbf{578}} & \bestcell{\textbf{$-68$\%}} & \bestcell{\textbf{64.43}} & \bestcell{\textbf{$+3.74$}}
                                & \bestcell{\textbf{864}} & \bestcell{\textbf{$-58$\%}} & \bestcell{\textbf{56.46}} & \bestcell{\textbf{$+4.84$}} \\
\bottomrule
\end{tabular}
}
\end{table}

\begin{table}[!htbp]
\caption{\textbf{Per-benchmark best-checkpoint detail.} \emph{Difficult-only S1} vs.\ \medpsy{} S2 on the smoke-eval suite. \medpsy{} S2 wins on the average and on every individual benchmark at both model sizes; the largest gap is on the hardest benchmark, MedXpertQA.}
\label{tab:abl-j-detail}
\centering
\footnotesize
\setlength{\tabcolsep}{3pt}
\adjustbox{max width=\linewidth}{%
\begin{tabular}{l c c c c c c c c}
\toprule
& \textbf{Best step} & \textbf{MedQA} & \textbf{MedXpertQA} & \textbf{PubMedQA} & \textbf{Avg} & \textbf{med-tok} & \textbf{mxq-tok} & \textbf{pmd-tok} \\
\midrule
\multicolumn{9}{l}{\emph{4B class}}\\
\emph{Difficult-only S1}                               &   4 & 82.09 & 26.94 & 76.20 & 61.74 & 1{,}744 & 2{,}756 & 620 \\
\textbf{\mqfourb{} S2} (step $200$)                    & 200 & \bestcell{84.84} & \bestcell{31.27} & \bestcell{77.20} & \bestcell{64.43} & \bestcell{641} & \bestcell{1{,}041} & \bestcell{343} \\
\midrule
\multicolumn{9}{l}{\emph{1.7B class}}\\
\emph{Difficult-only S1}                               &   4 & 65.99 & 16.49 & 69.20 & 50.56 & 2{,}088 & 2{,}758 & 704 \\
\textbf{\mqonesevenb{} S2} (step $330$)                & 330 & \bestcell{74.86} & \bestcell{21.31} & \bestcell{73.20} & \bestcell{56.46} & \bestcell{1{,}020} & \bestcell{1{,}251} & \bestcell{534} \\
\bottomrule
\end{tabular}
}
\end{table}

\subsubsection{Limitations of these ablations}
\label{subsec:abl-limitations}
Three caveats apply. First, all numbers in this appendix are computed on the 3-benchmark smoke-eval suite (MedQA, MedXpertQA, PubMedQA) under the \texttt{inspect-ai} flexible scorer; the per-ablation deltas reproduce under the LLM-as-Parser scorer used for the released checkpoints (Sec.~\ref{subsec:eval-method}), but absolute scores differ slightly between the two scorers due to different prompting conventions and post-generation handling. Second, no single ablation in this appendix trains for substantially more compute than the \medpsy{} component it replaces; we cannot rule out that with $5$--$10\times$ more steps the gap between mined and random Stage-2, or between \medpsy{} and \emph{No overlong}, would close further than reported. Third, the \emph{Difficult-only S1} ablation trains on a much smaller and less diverse corpus than \medpsy{} ($1{,}276$ items at 4B and $1{,}888$ at 1.7B vs.\ $\sim 15$--$16$K items in \medpsy{} Stage 1). We compensate via $50$ vs.\ $4$ epochs (4B) and $30$ vs.\ $4$ epochs (1.7B) to match the \medpsy{} Stage-1 step count; nevertheless, the diversity of the training distribution remains a confound, and a fully matched-data-volume single-stage comparison is left to future work.

\subsection{Existing assets, licenses, and terms of use}
\label{app:asset-licenses}

Tables~\ref{tab:asset-licenses-data}, \ref{tab:asset-licenses-models} and \ref{tab:asset-licenses-software} report the licenses or governing terms for the existing datasets, benchmarks, model checkpoints, and software tools used in this paper.

\begin{table}[p]
\caption{\textbf{Existing data and benchmark assets.} Licenses are listed for the source release or dataset card used in our experiments.}
\label{tab:asset-licenses-data}
\centering
\scriptsize
\setlength{\tabcolsep}{3pt}
\renewcommand{\arraystretch}{1.15}
\adjustbox{max width=\linewidth}{%
\begin{tabular}{p{0.17\textwidth} p{0.26\textwidth} p{0.24\textwidth} p{0.27\textwidth}}
\toprule
\textbf{Asset type} & \textbf{Asset / source} & \textbf{License / terms} & \textbf{Use in this paper} \\
\midrule
Training seed data & Genesis III biology/medicine subsets~\cite{genesis3} & CC-BY-NC-4.0 & Prompt-side question/option seeds for Corpus~1. \\
Training seed data & FineFineWeb health subset ~\cite{finefineweb} & Apache-2.0 & Health-domain passages used to generate additional MCQ prompt seeds. \\
Filtering model/tool & Ultra-FineWeb-classifier~\cite{ultrafineweb} & Apache-2.0 & Filtering FineFineWeb health-domain passages before prompt generation. \\
Focused SFT source / benchmark & AfriMedQA ~\cite{afrimedqa} & CC-BY-4.0 & Focused clinical QA prompt source for Corpus~2 and closed-ended evaluation benchmark. \\
Focused SFT source / benchmark & MedMCQA~\cite{medmcqa} & Apache-2.0 & Focused clinical QA prompt source for Corpus~2 and closed-ended evaluation benchmark. \\
Focused SFT source / benchmark & MedQA-USMLE~\cite{medqa} & CC-BY-4.0 & Focused clinical QA prompt source for Corpus~2 and closed-ended evaluation benchmark. \\
RL prompt source & AlphaMed19K~\cite{alphamedqa} & MIT & RL prompt pool for DAPO training and ablations. \\
Evaluation benchmark & MMLU health subdomains~\cite{mmlu} & MIT & Closed-ended health-literacy evaluation. \\
Evaluation benchmark & MMLU-Pro Health~\cite{mmlu-pro} & MIT & Closed-ended health-literacy evaluation. \\
Evaluation benchmark & MedXpertQA~\cite{medxpertqa} & MIT & Closed-ended expert-reasoning evaluation. \\
Evaluation benchmark & PubMedQA~\cite{pubmedqa} & MIT & Closed-ended biomedical-research evaluation. \\
Evaluation benchmark & HealthBench and HealthBench-Hard~\cite{healthbench} & MIT & Open-ended clinical-scenario evaluation. \\
Evaluation benchmark & LongHealth~\cite{adams2024longhealth} & Apache-2.0 & Long-context clinical-document evaluation (App.~\ref{app:longhealth}). \\
\bottomrule
\end{tabular}}
\end{table}

\begin{table}[p]
\caption{\textbf{Existing model assets used as backbones, teachers, graders, parsers, and baselines.} Custom model licenses and acceptable-use policies are reported by name where applicable.}
\label{tab:asset-licenses-models}
\centering
\scriptsize
\setlength{\tabcolsep}{3pt}
\renewcommand{\arraystretch}{1.15}
\adjustbox{max width=\linewidth}{%
\begin{tabular}{p{0.17\textwidth} p{0.27\textwidth} p{0.24\textwidth} p{0.26\textwidth}}
\toprule
\textbf{Asset type} & \textbf{Asset / source} & \textbf{License / terms} & \textbf{Use in this paper} \\
\midrule
Student backbone / baseline & Qwen3-4B-Thinking-2507 / Qwen3-4B family~\cite{qwen3} & Apache-2.0 & 4B student backbone and general baseline. \\
Student backbone / baseline & Qwen3-1.7B (Thinking)~\cite{qwen3} & Apache-2.0 & 1.7B student backbone and general baseline. \\
SFT target teacher & Baichuan-M3-235B~\cite{baichuanm3} & Apache-2.0 & Assistant-target generation and teacher-selection candidate. \\
Prompt generator / teacher candidate & Fleming-R1-32B~\cite{fleming-r1} & Apache-2.0 & MCQ generation from FineFineWeb health passages and teacher-selection candidate. \\
Teacher candidate / grader / parser & GPT-OSS-120B~\cite{gptoss} & Apache-2.0 & Teacher-selection candidate, HealthBench grader, and parser-robustness model. \\
Closed-ended parser & QwQ-32B~\cite{qwq} & Apache-2.0 & Main LLM-as-parser model for closed-ended answer extraction. \\
Grader / parser & Llama-3.3-70B-Instruct~\cite{llama33} & Llama 3.3 Community License & HealthBench grader and parser-robustness model. \\
HealthBench grader & CompassJudger-2-32B-Instruct~\cite{compassjudger} & Apache-2.0 & Main HealthBench grader and per-dimension grader. \\
\midrule
Medical baseline & MedGemma-27B-text-it ~\cite{medgemma2025} & Health AI Developer Foundations License / Google Health AI terms & Large medical baseline. \\
Medical baseline & MedGemma-1.5-4B-it~\cite{medgemma15} & Health AI Developer Foundations License / Google Health AI terms & Compact medical baseline. \\
Backbone-selection baseline & Qwen3-4B (Thinking)~\cite{qwen3} & Apache-2.0 & 4B-class backbone-selection baseline. \\
Backbone-selection baseline & Llama-3.2-3B-Instruct~\cite{llama32} & Llama 3.2 Community License & 4B-class backbone-selection baseline. \\
Backbone-selection baseline & Llama-3.2-1B-Instruct~\cite{llama32} & Llama 3.2 Community License & Sub-2B backbone-selection baseline. \\
Backbone-selection baseline & gemma-3-4b-it & Gemma Terms of Use & 4B-class backbone-selection baseline. \\
Backbone-selection baseline & SmolLM3-3B & Apache-2.0 & 4B-class backbone-selection baseline. \\
Backbone-selection baseline & SmolLM2-1.7B-Instruct & Apache-2.0 & Sub-2B backbone-selection baseline. \\
Backbone-selection baseline & LFM2.5-1.2B-Thinking & LFM1.0 custom model license & Sub-2B backbone-selection baseline. \\
Transfer-ablation backbone & DeepSeek-R1-Distill-Qwen-1.5B~\cite{deepseekr1} & MIT & Qwen2.5-derived reasoning backbone (Qwen2.5-Math-1.5B base, Apache-2.0) for the recipe-transfer experiment (App.~\ref{app:backbone-transfer}). \\
Large medical baseline & Meditron3-70B~\cite{meditron} & Llama 3.1 Community License & Additional large medical baseline. \\
Medical baseline & Meditron3-8B~\cite{meditron} & Llama 3.1 Community License & Additional medical baseline. \\
Large medical baseline & OpenBioLLM-70B~\cite{openbiollm} & Llama 3 license & Additional large medical baseline. \\
Medical baseline & JSL-MedLlama-3-8B-v2.0~\cite{medllama} & CC-BY-NC-ND-4.0 & Additional medical baseline. \\
\bottomrule
\end{tabular}}
\end{table}

\begin{table}[p]
\caption{\textbf{Existing software and infrastructure assets.} Commercial or hosted services are listed with their governing service terms.}
\label{tab:asset-licenses-software}
\centering
\scriptsize
\setlength{\tabcolsep}{3pt}
\renewcommand{\arraystretch}{1.15}
\adjustbox{max width=\linewidth}{%
\begin{tabular}{p{0.18\textwidth} p{0.27\textwidth} p{0.24\textwidth} p{0.25\textwidth}}
\toprule
\textbf{Asset type} & \textbf{Asset / source} & \textbf{License / terms} & \textbf{Use in this paper} \\
\midrule
Evaluation software & Inspect AI~\cite{inspectai} & MIT & Lightweight smoke-eval framework for RL ablations. \\
Evaluation software & OpenCompass~\cite{opencompass} & Apache-2.0 & LLM-as-Parser and model-grading evaluation codebase for the main benchmark suite. \\
Evaluation software & Medmarks v0.1.0~\cite{medmarks} & MIT & Matched-instance pairwise win-rate computation (App.~\ref{app:pairwise-winrate}). \\
Inference / quantization software & llama.cpp~\cite{llamacpp} & MIT & GGUF conversion, quantization, and evaluation of quantized variants. \\
Training framework & NVIDIA NeMo 25.09 & Apache-2.0 for NeMo framework& Distributed SFT training stack. \\
Training framework & verl~\cite{verl} & Apache-2.0 & Distributed RL training stack. \\
Training / inference runtime & PyTorch 2.5.1 & BSD-style license & Model training and evaluation runtime. \\
Inference engine & vLLM & Apache-2.0 & RL rollout and inference serving. \\
Experiment tracking & Weights \& Biases & MIT  & Experiment tracking and run logging. \\
Experiment tracking & TensorBoard & Apache-2.0 & Training visualization and logs. \\
Container runtime & Enroot & Apache-2.0 & Container runtime for training environment. \\
Cluster scheduler & SLURM & GPLv2-or-later & Cluster job scheduling. \\
GPU software stack & CUDA 12.1 / NVIDIA drivers & NVIDIA CUDA Toolkit EULA and NVIDIA driver terms & GPU training and inference software stack. \\
\bottomrule
\end{tabular}}
\end{table}

\let\subsection\oldsubsection

\end{document}